\documentclass[11pt,x11names]{article}
\usepackage{setspace}
\PassOptionsToPackage{table}{xcolor}
\usepackage{misc/emnlp2023}
\usepackage{times}
\usepackage{graphicx}
\usepackage{xfrac} 
\usepackage[T1]{fontenc}
\usepackage[utf8]{inputenc}
\usepackage{microtype}
\usepackage{booktabs}
\usepackage{todonotes}
\usepackage{amsmath,amssymb,bbm}
\usepackage{cleveref}
\usepackage{xspace}
\usepackage{listings,algorithm, algorithmicx, algpseudocode}
\usepackage{soul} 
\usepackage[hypcap=false]{caption}
\usepackage{multirow,multicol}
\usepackage[normalem]{ulem}
\usepackage{float}
\usepackage{array}
\usepackage{bm}
\usepackage{xfrac}
\usepackage[shortlabels]{enumitem}
\usepackage{mathtools}
\usepackage{url} 
\usepackage{tcolorbox}
\usepackage{makecell} 
\usepackage{anyfontsize} 
\usepackage{footmisc}
\usepackage{arydshln} 
\usepackage{booktabs, multirow, colortbl, subcaption}
\usepackage{devanagari} 
\usepackage{enumitem}

\newenvironment{blue}{\color{blue}}{}
\newcommand\remove{\bgroup\markoverwith{\textcolor{orange}{\rule[.5ex]{2pt}{1pt}}}\ULon}

\setlist[itemize,enumerate]{noitemsep}

\usepackage{tabularx}

\crefname{lstlisting}{listing}{listings}
\Crefname{lstlisting}{Listing}{Listings}
\crefname{myequation}{equations}{equations}
\Crefname{myequation}{Equations}{Equations}
\Crefname{algorithmCaption}{Algorithm}{Algorithms}
\crefname{example}{example}{examples}
\Crefname{example}{Example}{Examples}
\crefname{prompt}{prompt}{prompts}
\Crefname{prompt}{Prompt}{Prompts}
\DeclareCaptionType{algorithmCaption}[Algorithm][List of algorithms]
\DeclareCaptionType{example}[Example][List of examples]
\DeclareCaptionType{prompt}[Prompt][List of prompts]
\DeclareCaptionType{myequation}[Equation][List of equations]

\usepackage{courier}
\usepackage{marginnote}

\definecolor{TodoColor}{rgb}{1,0.7,0.6}

\newcommand\cometpolycand{COMET\textsubscript{poly-cand}\xspace}
\newcommand\cometpolyic{COMET\textsubscript{poly-ic}\xspace}
\newcommand\gembapolycand{GEMBA\textsubscript{poly-cand}\xspace}
\newcommand\gembapolyic{GEMBA\textsubscript{poly-ic}\xspace}
\newcommand\knnpolycand{$k$-NN\textsubscript{poly-cand}\xspace}
\newcommand\knnpolyic{$k$-NN\textsubscript{poly-ic}\xspace}

\definecolor{FindingsColor}{gray}{0.85}

\makeatletter\def\Hy@Warning#1{}\makeatother
\let\svthefootnote\thefootnote
\newcommand\blankfootnote[1]{%
  \let\thefootnote\relax\footnotetext{#1}%
  \let\thefootnote\svthefootnote%
}

\author{%
Maike Züfle$^{1\,\bigstar}$ \quad Vilém Zouhar$^{2\,\bigstar}$ \quad Tu Anh Dinh$^{1\,\bigstar}$ \quad Felipe Maia Polo$^3$ \\
\bf Jan Niehues$^1$ \quad Mrinmaya Sachan$^2$ \\[0.7em]
$^1$Karlsruhe Institute of Technology \quad
$^2$ETH Zurich \quad
$^3$University of Michigan \\
\tt\small
\{\href{mailto:maike.zuefle@kit.edu}{\color{black} maike.zuefle},\href{mailto:tu.dinh@kit.edu}{\color{black} tu.dinh}\}@kit.edu
\,\,
\href{mailto:vzouhar.ethz.ch}{\color{black} vzouhar.ethz.ch}
}

\title{COMET-poly: Machine Translation Metric Grounded in Other Candidates}

\begin{document}

\maketitle

\blankfootnote{
\hspace{-1.5mm}$^\bigstar$Equal contribution, sorted anti-alphabetically.
}

\begin{abstract}
Automated metrics for machine translation attempt to replicate human judgment.
Unlike humans, who often assess a translation in the context of multiple alternatives, these metrics typically consider only the source sentence and a single translation.
This discrepancy in the evaluation setup may negatively impact the performance of automated metrics.
We propose two automated metrics that incorporate additional information beyond the single translation.
\cometpolycand uses alternative translations of the same source sentence to compare and contrast with the translation at hand, thereby providing a more informed assessment of its quality.
\cometpolyic, inspired by retrieval-based in-context learning, takes in translations of similar source texts along with their human-labeled quality scores to guide the evaluation.
We find that including a single additional translation in \cometpolycand improves the segment-level metric performance (0.079$\rightarrow$0.118 $\tau_b$), with further gains when more translations are added. Incorporating retrieved examples in \cometpolyic yields similar improvements  (0.079$\rightarrow$0.116 $\tau_b$). We release our models publicly.\footnote{We release the paper \href{https://github.com/zouharvi/COMET-multi-cand/}{code} and pre-trained quality estimation models \href{https://huggingface.co/collections/zouharvi/comet-poly-687036a351c42f3ef89ebcce}{\cometpolyic} and \href{https://huggingface.co/collections/zouharvi/comet-poly-687036a351c42f3ef89ebcce}{\cometpolycand}.}
\end{abstract}


\begin{figure}[t]
\includegraphics[width=\linewidth]{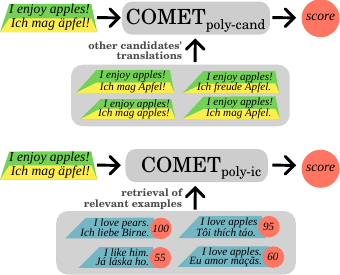}
\caption{
The \cometpolyic model consults a knowledge base of previously human-scored translations before assigning the quality estimation score to the candidate translation.
The \cometpolycand considers other possible translations apart from the candidate one.
Both metrics work better than just providing the source and the translation.}
\label{fig:highlevel_model_poly}

\vspace{-3mm}
\end{figure}

\section{Introduction}

There is a gap between how humans and automated metrics score translations.
Automated metrics receive the source segment, usually a sentence or a paragraph, a single translation, and optionally a reference translation.
They are then tasked with assessing the quality of the translation.
In contrast, human evaluation is less episodic.
Human raters often assess multiple translations in sequence \citep{graham-etal-2013-continuous,freitag-etal-2021-experts,kocmi-etal-2024-error}, considering them side-by-side.
Even though annotations are made for each translation individually, annotators become calibrated (known as sequence effect, \citealp{mathur-etal-2017-sequence}), to common error patterns and their own evaluation criteria as they review multiple translations.
As a result, they effectively score each translation in the context of others. 
Moreover, unlike human annotators, who have a deep understanding of the languages involved and can assess a wide range of translation qualities, automated metrics are limited by the data they were trained on. As a result, their performance tends to degrade when evaluating translations that deviate from their training distributions, such as out-of-domain content \citep{zouhar-etal-2024-fine}.

We present two conceptual approaches to address these two challenges by incorporating additional context into standard automated metrics, such as COMET \citep{rei-etal-2020-comet}. Our main motivation is to narrow the gap between human evaluation and automated metrics, enabling automated metrics to score translations in the context of other translations
and making them more robust to out-of-domain data. Specifically, we introduce two models trained within this framework, \cometpolycand and \cometpolyic:
\vspace{-4pt}
\begin{itemize}[leftmargin=*, labelsep=-3pt, align=left]
    \item ~~ In \cometpolycand, different translations of the same source sentence are provided to the model as additional context (\Cref{fig:highlevel_model_poly} top). This is suitable for scenarios such as (1) benchmarking, where we evaluate translations of multiple systems on the same source sentence, or (2) reranking, where we need to select the best translation from a pool of candidate translations.

    \item ~~ In \cometpolyic, which is inspired by retrieval-based in-context learning, tuples of (\textit{source}, \textit{translation}, \textit{human quality score}) are provided to the model as additional context. The tuples are retrieved based on the source sentence similarity to the evaluation example at hand (\Cref{fig:highlevel_model_poly} bottom). In practice, in-context examples can be obtained from existing, previously scored translations—such as those found in prior WMT annotation datasets \citep{freitag-etal-2024-llms,kocmi-etal-2024-findings}.
\end{itemize}

This paper is structured as follows.
In  \Cref{sec:methods}, we first describe the task of machine translation quality estimation (QE) and COMET \citep{rei-etal-2020-comet}, a popular QE metric.
Then, we describe our two proposed model variants, \cometpolycand and \cometpolyic.
We also apply the same approach to two additional QE systems with contrasting characteristics: a non-parametric $k$-NN baseline and GEMBA, a large parametric LLM-based evaluator.
In \Cref{sec:setup,sec:results}, we show that our methods not only improve COMET’s segment-level performance but also outperform both the much larger GEMBA model and the $k$-NN baseline, despite their simplicity.
These approaches also show promise for instant on-the-fly domain adaptation.
We place our contributions in context with related work in \Cref{sec:related_work}.
Finally, in \Cref{sec:conclusion},  we provide some practical guidance on using these metrics along with potential caveats.

We publicly release our models under open license, and submit our models to the \href{https://www2.statmt.org/wmt25/mteval-subtask.html}{WMT 2025 Metrics Shared Task}.

\section{Methods}
\label{sec:methods}

In this section, we introduce the translation quality estimation task, review COMET, and present two extensions for improved quality estimation and domain adaptation.

\subsection{Background}

\paragraph{Quality estimation (QE).} Given a source text $s$ and a model-produced translation (MT) $t$, which is assessed by a human annotator on a scale from 0\% to 100\%, the goal of quality estimation (QE) is to develop a metric to predict this score.

\paragraph{Baseline COMET.}  Traditionally, quality estimation relied on static, rule-based metrics, but the field has shifted toward learned, data-driven metrics that can better approximate human judgments \cite{freitag-etal-2022-results}.
Learned automated metrics can be thought of as a function $f$, taking a source sentence $s$ and a translation $t$ as input and producing a continuous score $f(s, t)\in[0,1]$. $f$ is usually trained in a supervised manner to approximate human judgment $y_{s,t}=\mathrm{human}(s, t)$:
\begin{align*}
f(s, t) \xrightarrow{\text{train}} y_{s,t} \label{eq:base_optimize}
\end{align*}
A popular recent choice for $f$ is 
COMET \citep{rei-etal-2020-comet}, which is a combination of a trainable encoder model $e_{\theta_1}$ and a multi-layer perception head $\mathrm{MLP}_{\theta_2}$. COMET first embeds the source and translation texts, obtaining $s^e=e_{\theta_1}(s)$ and $t^e=e_{\theta_1}(t)$, and then transforms the embeddings into a score prediction using $\mathrm{MLP}_{\theta_2}$.
We denote the set of trainable weights as $\theta = (\theta_1,\theta_2)$.
Specifically, COMET is formulated as:
\begin{align*}
\mathrm{COMET}_\theta(s, t) = \mathrm{MLP}_{\theta_2}\big(g_{\theta_1}(s,t)\big),\\
\text{with }g_{\theta_1}(s,t) = \langle s^e, t^e, |s^e - t^e|, s^e*t^e \rangle.
\end{align*}
Here, $g_{\theta_1}$ constructs a feature vector for the pair $(s, t)$ by concatenating their embeddings $s^e$ and $t^e$ with additional element-wise transformations: the absolute difference $\lvert s^e - t^e \rvert$ and the element-wise product $s^e * t^e$.
The trainable weights $\theta$ are optimized by minimizing the mean squared error between the COMET score and human labels using a variation of the stochastic gradient descent algorithm.

While the baseline COMET framework is effective, it does not support incorporating additional information. 
Just like for human evaluators, having more information such as (1) multiple candidates' translations for the same source, or (2) ground-truth example quality scores of translations, could improve the performance of COMET further.
Thus, we introduce two extensions for COMET, which we train from scratch..

\subsection{Multiple Candidates: \cometpolycand}
\label{sec:comet_poly_cand}
Our first variant targets scenarios like benchmarking or reranking MT models, where multiple translations of the same source segment are available. In these cases, we extend the model’s context by including additional translations $\{t_i\}_{i=2}^n$ of the same source sentence $s$, allowing the model to leverage multiple candidate translations simultaneously. \Cref{fig:highlevel_model_poly} (top) shows an illustration of this model architecture. 

Specifically, we include the embeddings of these additional translations as part of the input to the multi-layer perceptron.
Formally, for all $i\in \{2,\cdots,n\}$, we define
\begin{align*}
g_{\theta_1}(t, t_i) = \langle t_i^e, |t_i^e - t^e|, t_i^e*t^e \rangle.
\end{align*}
We then concatenate $\langle g_{\theta_1}(s, t), g_{\theta_1}(t, t_1),...,$ $g_{\theta_1}(t, t_n) \rangle$ and pass it to the MLP. During training, we ensure that the additional translations $\{t_i\}_{i=2}^n$ differ from the main translation $t$, and keep $n$ fixed across all training examples. 

\paragraph{Joint predictions.} To reduce computation time, \cometpolycand can be trained to jointly predict the quality scores of the original translation along with those of the additional translations. The training objective then becomes:
\begin{align*}
f(s, t, t_2, ..., t_n) \xrightarrow{\text{train}} y_{s,t}, y_{s,t_2},...,y_{s,t_n}
\end{align*}

\paragraph{Using scores of other translations.}
When the human assessment scores for additional translations, $\{y_{s,t_i}\}_{i=2}^n$, are available, we can further augment the feature vector using these scores. The input to the MLP would become:
$$
\langle g_{\theta_1}(s, t), g_{\theta_1}(t, t_1), y_{s,t_1},...,g_{\theta_1}(t, t_n), y_{s,t_n} \rangle
$$
This is particularly useful when we wish to evaluate a new system on a pre-existing benchmark with other candidate translations whose qualities are already annotated by humans.

\subsection{In-context Learning: \cometpolyic}\label{sec:ic}

In the previous approach, we used additional translations of the same source sentence, a setup that might be unrealistic outside controlled scenarios such as benchmarking or reranking. An alternative, inspired by the success of \textit{in-context} learning in other domains \cite{brown2020language}, is to provide the model with other, similar examples: by conditioning on human-scored translations, it can learn the mapping between translation patterns and quality judgments on the fly. 

\cometpolyic implements this by retrieving source–translation–score triplets from a knowledge base, in our case, prior WMT annotation datasets \citep{freitag-etal-2024-llms,kocmi-etal-2024-findings}, and using them as context, enabling the model to adapt its evaluation to different domains.
An illustration is shown in \Cref{fig:highlevel_model_poly} (bottom).

Specifically, for each input example (source $s$, translation $t$), we retrieve the examples $\{(s_i,t_i,y_{s_i,t_i})\}_{i=2}^{n_\text{ICL}+1}$ from a knowledge base $\mathcal{D}$.
The new examples are added to the representation vector similar to \cometpolycand, considering both embeddings and labels, by appending
\begin{align*}
    &\langle t_i^e, |t_i^e - t^e|, t_i^e*t^e,s_i^e, |s_i^e - s^e|, |s_i^e * s^e|, y_{s_i,t_i} \rangle
\end{align*}
to $g_{\theta_1}(s,t)$ for all $i\in \{2,\cdots,n_\text{ICL}+1\}$.

The ICL examples are retrieved using normalized embedding (cosine) similarity computed from either the source $s^e$ (default), the translation $t^e_i$, their arithmetic combination $s^e+t^e_i$, or their concatenation $\langle s^e, t^e_i \rangle$.
We retrieve up to five most similar examples, discarding exact matches during training.
We present detailed ablations of different filtering and retrieval setups in \Cref{sec:results}.

\subsection{Including Reference Translations}
Optionally, COMET can also make use of a reference translation \citep{rei-etal-2020-comet}, though this is no longer part of the standard QE setup.
We also report results for \cometpolycand and \cometpolyic in the reference-based setting, by incorporating the reference $r$ in their inputs, i.e., $f(s, t, r)$. However, our primary focus remains on QE, as references are often unavailable in practical scenarios.

\subsection{Models Beyond COMET}\label{sec:beyond}

Since our method is not specific to COMET, we include two models that, like our extensions, can take multiple candidate translations into account.

\paragraph{$\bm{k}$-nearest neighbors.}
As our first baseline method, we propose using a $k$-nearest-neighbours ($k$-NN) approach, mirroring methods used in similar contexts \citep{dinh-etal-2024-quality-fixed}. $k$-NN naturally leverages existing high-quality examples by retrieving similar instances, providing a strong non-parametric baseline that complements our model-based approaches. The $k$-NN baseline is implemented for our two different setups: \knnpolycand and \knnpolyic. 

For \knnpolycand and for a pair $(s,t)$, we retrieve $k$ additional translations for $s$. These are selected based on the cosine similarity of the target translation $t$ and candidate translations, where embeddings are obtained using the
 \href{https://huggingface.co/sentence-transformers/all-MiniLM-L12-v2}{all-MiniLM-L12-v2} \citep{reimers-gurevych-2020-making}, yielding the set $\{t_i\}_{i=2}^{k+1}$. We then rate each candidate using Baseline COMET, obtaining $\{\text{COMET}(s,t_i)\}_{i=2}^{k+1}$. Finally, we use their average as the final prediction, i.e.,
\[
\hat{y}^{\text{\knnpolycand}}_{s,t} = \frac1k\sum_{i=2}^{k+1} \text{COMET}(s,t_i).
\]
For \knnpolyic, we retrieve $k=n_\text{ICL}$ examples, $\{(s_i,t_i,y_{s_i,t_i})\}_{i=2}^{k+1}$, following the retrieval strategies described in \Cref{sec:ic}, and then average their human scores to obtain the prediction:
\[
\hat{y}^{\text{\knnpolyic}}_{s,t} = \frac{1}{k}\sum_{i=2}^{k+1} y_{s_i,t_i}.
\]
We further extend the $k$-NN approaches using weighted averages in Appendix \ref{append:knn}.

\paragraph{Using LLMs as evaluators.}
As a second baseline, we use large language models (LLMs) for MT evaluation, leveraging their effectiveness in this task \citep{kocmi-federmann-2023-large}.
Specifically, we apply in-context learning \citep{brown2020language}, a standard method for injecting new knowledge into LLMs at inference time. Similar to COMET\textsubscript{poly}, we provide LLMs with additional contextual information when scoring translations. However, unlike COMET\textsubscript{poly} variants, which update model parameters during training, LLMs receive this information only through their prompts at inference time, without any parameter modification.

For prompt creation, we build on top of GEMBA \citep{kocmi-federmann-2023-large}, a framework designed to prompt LLMs to score the quality of translations. Leveraging GEMBA’s pre-defined prompts, we extend them to two settings:  (1) \gembapolycand, where additional translations of the same source sentence are provided, and (2) \gembapolyic, where full examples (including source, translation, and human quality score) are included. Prompt details are provided in Appendix~\ref{sec:implementation_details_gemba}.

\section{Experimental Setup}
\label{sec:setup}


This section outlines the training and evaluation procedures, as well as the experimental setup.

\paragraph{Data.}
We use the direct assessment scorings of WMT up to 2023 (inclusive) for training (600k segments).
For testing and evaluation, we use WMT 2024 (105 segments), which has been evaluated with the ESA protocol \citep{kocmi-etal-2024-error}.
This dataset covers eleven language pairs: English to Czech, German, Spanish, Hindi, Icelandic, Japanese, Russian, Chinese, Czech to Ukrainian, and Japanese to Chinese.
From ESA, we use the final scores (as opposed to error spans), which have the same scale as direct assessment.
For MQM, we convert the error span annotations on a translation to the final score by taking $ 1-(5 \cdot major + 1 \cdot minor)/100 $, where $major$ is the number of annotated major errors, and $minor$ is the number of minor errors annotated in the translation. In this way, the scores are aligned roughly on the same scale compared to DA scores.

\begin{table*}[t]
\small
\centering
\begin{tabular}{r>{\hspace{-2mm}}lccc@{\hspace{8mm}}ccc>{\color{gray}\hspace{-2mm}}l}
\cmidrule[1pt]{2-8}
& & \multicolumn{3}{c}{\bf Reference-less} & \multicolumn{3}{c}{\bf Reference-based} \\
& \bf Model
& $\bm{\rho \uparrow}$ & $\bm{\tau_b \uparrow}$ & \hspace{-2mm}$\mathrm{\textbf{MAE}}\bm{\downarrow}$\hspace{-2mm}
& $\bm{\rho \uparrow}$ & $\bm{\tau_b \uparrow}$ & \hspace{-2mm}$\mathrm{\textbf{MAE}}\bm{\downarrow}$\hspace{-2mm} & \\
\cmidrule{2-8}
standard COMET model & $f(s, t) \rightarrow \hat{y_t}$ 
& \cellcolor{Goldenrod3!0.000} 0.105 & \cellcolor{Chartreuse3!0.000} 0.079 & \cellcolor{OrangeRed2!100.000} 30.2 
& \cellcolor{Goldenrod3!52.045} 0.245 & \cellcolor{Chartreuse3!51.786} 0.166 & \cellcolor{OrangeRed2!75.839} 26.6 
\\[-0.1em]

& \bf \cometpolycand \\
additional candidate & $f(s, t, t^*_2) \rightarrow \hat{y_t}$ 
& \cellcolor{Goldenrod3!20.446} 0.160 & \cellcolor{Chartreuse3!28.571} 0.127 & \cellcolor{OrangeRed2!88.591} 28.5 
& \cellcolor{Goldenrod3!65.428} 0.281 & \cellcolor{Chartreuse3!60.119} 0.180 & \cellcolor{OrangeRed2!73.826} 26.3 
\\
additional candidate, output joint predictions & $f(s, t, t^*_2) \rightarrow \hat{y_t}, \hat{y_{t^*_2}}$ 
& \cellcolor{Goldenrod3!23.048} 0.167 & \cellcolor{Chartreuse3!20.238} 0.113 & \cellcolor{OrangeRed2!90.604} 28.8 
& \cellcolor{Goldenrod3!63.197} 0.275 & \cellcolor{Chartreuse3!55.357} 0.172 & \cellcolor{OrangeRed2!69.128} 25.6 
\\
additional candidate and its score & $f(s, t, t^*_2, y_{t^*2}) \rightarrow \hat{y_t}$ 
& \cellcolor{Goldenrod3!60.223} 0.267 & \cellcolor{Chartreuse3!76.190} 0.207 & \cellcolor{OrangeRed2!44.295} 21.9 
& \cellcolor{Goldenrod3!100.000} 0.374 & \cellcolor{Chartreuse3!97.619} 0.243 & \cellcolor{OrangeRed2!35.570} 20.6 
\\

& \bf \cometpolyic \\
additional candidate and its score  & $f(s, t, t^*_2, y_{t^*2}) \rightarrow \hat{y_t}$  
& \cellcolor{Goldenrod3!13.383} 0.141 & \cellcolor{Chartreuse3!0.000} 0.116 & \cellcolor{OrangeRed2!100.000} 27.3 
& \cellcolor{Goldenrod3!91.822} 0.352 & \cellcolor{Chartreuse3!100.000} 0.247 & \cellcolor{OrangeRed2!0.000} 15.3  
\\\\[-0.9em]
\cmidrule[1pt]{2-8}
\end{tabular}
\caption{Results for \cometpolycand and \cometpolyic. The first row shows the standard COMET. The middle and bottom parts show that adding additional translation candidates and in-context examples boosts performance.}
\label{tab:macarons_cream_kuchen}
\bigskip
\end{table*}

\begin{table*}[t]
\small
\centering
\setlength{\tabcolsep}{2pt}
\begin{tabular}{
    l
    ccccc
    ccccc
    ccccc
}
\toprule
\bf Model (Reference-less)\hspace{-1cm}
  & \multicolumn{5}{c}{$\bm{\rho \uparrow}$}
  & \multicolumn{5}{c}{$\bm{\tau_b \uparrow}$}
  & \multicolumn{5}{c}{$\mathrm{\textbf{MAE}}\bm{\downarrow}$} \\
(+additional)
  & +1 & +2 & +3 & +4 & +5
  & +1 & +2 & +3 & +4 & +5
  & +1 & +2 & +3 & +4 & +5 \\
\midrule
$f(s, t) \rightarrow \hat{y_t}$
  & \cellcolor{Goldenrod3!15.789} 0.105
  & \cellcolor{Goldenrod3!15.789} 0.105
  & \cellcolor{Goldenrod3!15.789} 0.105
  & \cellcolor{Goldenrod3!15.789} 0.105
  & \cellcolor{Goldenrod3!15.789} 0.105
  & \cellcolor{Chartreuse3!0.000} 0.079
  & \cellcolor{Chartreuse3!0.000} 0.079
  & \cellcolor{Chartreuse3!0.000} 0.079
  & \cellcolor{Chartreuse3!0.000} 0.079
  & \cellcolor{Chartreuse3!0.000} 0.079
  & \cellcolor{OrangeRed2!100.000} 30.2
  & \cellcolor{OrangeRed2!100.000} 30.2
  & \cellcolor{OrangeRed2!100.000} 30.2
  & \cellcolor{OrangeRed2!100.000} 30.2
  & \cellcolor{OrangeRed2!100.000} 30.2 \\\\[-0.2em]

\textbf{\cometpolycand} \\  
$f(s, t, t_{\cdots}) \rightarrow \hat{y_t}$
  & \cellcolor{Goldenrod3!36.842} 0.160
  & \cellcolor{Goldenrod3!68.421} 0.251
  & \cellcolor{Goldenrod3!57.895} 0.224
  & \cellcolor{Goldenrod3!52.632} 0.202
  & \cellcolor{Goldenrod3!47.368} 0.190
  & \cellcolor{Chartreuse3!31.579} 0.127
  & \cellcolor{Chartreuse3!42.105} 0.145
  & \cellcolor{Chartreuse3!31.579} 0.127
  & \cellcolor{Chartreuse3!31.579} 0.130
  & \cellcolor{Chartreuse3!26.316} 0.120
  & \cellcolor{OrangeRed2!89.474} 28.5
  & \cellcolor{OrangeRed2!84.211} 27.6
  & \cellcolor{OrangeRed2!78.947} 26.8
  & \cellcolor{OrangeRed2!89.474} 28.4
  & \cellcolor{OrangeRed2!84.211} 27.6 \\

$f(s, t, t_{\cdots}, y_{t_{\cdots}}) \rightarrow \hat{y_t}$
  & \cellcolor{Goldenrod3!78.947} 0.267
  & \cellcolor{Goldenrod3!94.737} 0.321
  & \cellcolor{Goldenrod3!100.000} 0.328
  & \cellcolor{Goldenrod3!100.000} 0.327
  & \cellcolor{Goldenrod3!94.737} 0.321
  & \cellcolor{Chartreuse3!84.211} 0.207
  & \cellcolor{Chartreuse3!94.737} 0.229
  & \cellcolor{Chartreuse3!94.737} 0.230
  & \cellcolor{Chartreuse3!100.000} 0.235
  & \cellcolor{Chartreuse3!100.000} 0.233
  & \cellcolor{OrangeRed2!47.368} 21.9
  & \cellcolor{OrangeRed2!21.053} 17.3
  & \cellcolor{OrangeRed2!15.789} 16.0
  & \cellcolor{OrangeRed2!0.000} 14.0
  & \cellcolor{OrangeRed2!0.000} 13.7 \\\\[-0.2em]

\textbf{\cometpolyic} \\  
$f(s, t, t_{\cdots}, y_{t_{\cdots}}) \rightarrow \hat{y_t}$
  & \cellcolor{Goldenrod3!26.316} 0.141
  & \cellcolor{Goldenrod3!26.316} 0.134
  & \cellcolor{Goldenrod3!31.579} 0.148
  & \cellcolor{Goldenrod3!21.053} 0.128
  & \cellcolor{Goldenrod3!0.000} 0.068
  & \cellcolor{Chartreuse3!26.316} 0.116
  & \cellcolor{Chartreuse3!21.053} 0.108
  & \cellcolor{Chartreuse3!26.316} 0.114
  & \cellcolor{Chartreuse3!21.053} 0.105
  & \cellcolor{Chartreuse3!0.000} 0.075
  & \cellcolor{OrangeRed2!84.211} 27.3
  & \cellcolor{OrangeRed2!84.211} 27.2
  & \cellcolor{OrangeRed2!68.421} 24.7
  & \cellcolor{OrangeRed2!84.211} 27.6
  & \cellcolor{OrangeRed2!84.211} 27.4 \\
\bottomrule
\end{tabular}

\caption{Results for \cometpolycand and \cometpolyic using different numbers of additional translation candidates. The +1 is equal to the results in \Cref{tab:macarons_cream_kuchen}. The +x uses x additional translation candidates, which improves performance especially for \cometpolycand.}
\label{tab:crisps}
\bigskip
\end{table*}

\paragraph{Training.}
We train the Baseline COMET model, \cometpolycand and \cometpolyic based on pretrained RoBERTa \citep{liu2019roberta} on WMT human judgment data for five epochs.
For \cometpolycand, we retrieve up to five candidate translations, either randomly or based on embedding similarity.
For \cometpolyic, we retrieve up to five in-context examples from the training data based on embedding similarity.
The metrics are trained in a maximally comparable model setup, which is detailed in \Cref{sec:implementation_details}.

\paragraph{Evaluation.}
We evaluate the metrics on the segment level in three ways: Pearson correlation, Kendall's tau-b, and Mean Absolute Error (MAE).
In contrast to \citet{freitag-etal-2024-llms} we do not do perform any group-by-item nor group-by-item.
Results are macro-averaged across eleven languages.

Pearson correlation measures the linear relationship between metric scores and human ratings: higher values indicate better alignment, though not necessarily on the same scale. Mean Absolute Error (MAE), in contrast, captures the average absolute difference between metric and human scores, with lower values indicating closer agreement in both value and scale. Kendall's tau-b focuses on rank correlation, reflecting how well the metric preserves the relative ordering of translations.
While Pearson and Kendall's tau-b range from -1 to 1, MAE is unbounded and depends on the scoring scale.

\paragraph{Experiments.}
To ensure a controlled evaluation setting, we first train a standard COMET model on the data described before and use it as a baseline. We then investigate \cometpolycand by incorporating additional translations into the base model and analysing the impact of different selection strategies. Similarly, we explore \cometpolyic, experimenting with various retrieval methods and assessing its potential for domain adaptation. We complement our experiments with \knnpolycand and \knnpolyic as non-parametric baselines, and \gembapolycand and \gembapolyic as large-parameter LLM baselines.

\section{Results and Analysis}
\label{sec:results}

In the following, we discuss and analyse the results of \cometpolycand and \cometpolyic, compare them to the non-parametric $k$-NN and the large parametric GEMBA model, and discuss the runtime impact of our method.

\subsection{Results for \cometpolycand}

\paragraph{Additional candidate helps.} We begin by evaluating \cometpolycand in its simplest setting:
adding a single additional translation from the same source as the candidate being scored. We choose the closest additional translation $t_2^*$ as, intuitively, the closer it is to the candidate $t$, the more relevant it is for assessing its quality.
We select $t_2^*$ based on the embedding distance computed between candidate translations (see \Cref{sec:implementation_details} for details on embeddings and distance metrics).
The corresponding results are shown in the middle part of \Cref{tab:macarons_cream_kuchen}. 

Across all evaluation metrics, including an additional translation $f(s, t, t_2)$, considerably improves performance compared to the standard COMET baseline $f(s, t)$. Specifically, Pearson correlation improved by over 50\%. 
The joint translation prediction objective, which scores both the original translation and the additional translation, also yields gains over the baseline, though it performs slightly worse than the single-prediction setup. This suggests that, in scenarios where faster inference is needed, the joint-prediction setup offers a practical trade-off, delivering improved performance with smaller additional cost.
Finally, including the gold score $y_{t_2}$ of the additional translation in the input vastly improves the metric performance. However, note that this is an ideal scenario where the gold score $y_{t_2}$ is available, which is not always realistic. 

Note that it is not always possible to find additional translations that are similar to the translation at hand. Therefore, we experiment with using a randomly selected additional candidate to test the robustness of \cometpolycand. This still results in notable gains, albeit smaller than with similar candidates.  We report these results in \Cref{app:polycand}.

\paragraph{More than one candidate helps.} We extend \cometpolycand by increasing the number of additional candidates. The results are shown in \Cref{tab:crisps}. Having more than one additional candidate further improves the performance of \cometpolycand, as we are providing a more global view of possible translations to the model. However, this effect starts to diminish beyond two additional candidates. For comparison, results using random additional candidates are provided in  \Cref{app:polycand}.

\paragraph{Additional translation complement reference.}
Previous experiments focused on reference-free evaluation. To complete the picture, we now explore how \cometpolycand performs when reference translations are available.

The right half of \Cref{tab:macarons_cream_kuchen} shows that using \cometpolycand with reference yields better performance than \cometpolycand in QE mode, though the gain is smaller than for standard COMET. 
This indicates that additional translations help narrow the gap but cannot fully replace references. Rather, additional translations complement references by providing further improvements on top of them.

\subsection{Results for \cometpolyic}
Building on this idea of leveraging additional context, we next evaluate \cometpolyic, which incorporates in-context examples to further enhance evaluation quality.

\paragraph{In-context examples help.} We retrieve an in-context example using the source text $s^e$, embedded via an external embedding model (details in \Cref{app:ic_ablations}).  Results in the bottom row of \Cref{tab:macarons_cream_kuchen} show that COMET benefits significantly from these examples, outperforming the baseline without in-context examples. This improvement also holds for \cometpolyic with references. However, compared to \cometpolycand, in-context examples appear less informative than additional candidates with the same source, resulting in slightly reduced performance.
We also test other embedding types (including COMET’s own) and variations using the target or both source and target for retrieval. However, none of these alternatives yields further improvements. Full ablations are presented in \Cref{app:ic_ablations}.

\paragraph{More in-context examples improve performance.} While a single in-context example already boosts performance, adding up to three examples leads to further improvements. As shown in the bottom half of \Cref{tab:crisps}, performance increases with the number of retrieved examples using the external embedding model and $s^e$ for retrieval, but declines beyond three examples, likely because additional examples become less similar and less relevant.

We also provide preliminary experiments in \Cref{app:ic_domain} on how \cometpolyic can leverage in-context examples to adapt its quality estimation to a new domain, and find a slight improvement compared to the base model.

\subsection{Adding Candidates to\linebreak Models Beyond COMET}

In order to see whether having additional candidates or examples also helps with other QE methods other than COMET, we look into the performance of two baselines: the non-parametric $\bm{k}$-nearest neighbors and large parametric LLM evaluator with GEMBA. 

We use $\bm{k}$-nearest neighbors in the retrieval setting for both \knnpolycand and \knnpolyic, i.e., retrieving similar examples along with their gold quality scores, since the gold scores are required for $\bm{k}$-nearest neighbors. 
For GEMBA, we experiment with all \gembapolycand variances (random/similar candidate, with/without gold scores) and \gembapolyic, similar to \cometpolycand and \cometpolyic.

\begin{table}[t]
    \small
    \centering
    \setlength{\tabcolsep}{2pt}
    \begin{subtable}[t]{0.49\columnwidth}
\centering
\caption{\knnpolycand}
\begin{tabular}{c c c c}
\toprule
$k$ & $\bm{\rho}\!\uparrow$ & $\bm{\tau_b}\!\uparrow$ & $\mathbf{MAE}\!\downarrow$\\
\midrule
1 & \cellcolor{Goldenrod3!93} 0.083 & \cellcolor{Chartreuse3!100} 0.064 & \cellcolor{OrangeRed2!78} 30.4\\
2 & \cellcolor{Goldenrod3!100} 0.087 & \cellcolor{Chartreuse3!100} 0.064 & \cellcolor{OrangeRed2!75} 30.3\\
3 & \cellcolor{Goldenrod3!98} 0.086 & \cellcolor{Chartreuse3!96} 0.062 & \cellcolor{OrangeRed2!78} 30.4\\
4 & \cellcolor{Goldenrod3!97} 0.085 & \cellcolor{Chartreuse3!90} 0.059 & \cellcolor{OrangeRed2!78} 30.4\\
5 & \cellcolor{Goldenrod3!97} 0.085 & \cellcolor{Chartreuse3!86} 0.057 & \cellcolor{OrangeRed2!78} 30.4\\
\bottomrule
\end{tabular}
\end{subtable}%
\hfill
\begin{subtable}[t]{0.49\columnwidth}
\centering
\caption{\knnpolyic}
\begin{tabular}{c c c c}
\toprule
$k$ & $\bm{\rho}\!\uparrow$ & $\bm{\tau_b}\!\uparrow$ & $\mathbf{MAE}\!\downarrow$\\
\midrule
1 & \cellcolor{Goldenrod3!0} 0.029 & \cellcolor{Chartreuse3!0} 0.014 & \cellcolor{OrangeRed2!100} 31.1\\
2 & \cellcolor{Goldenrod3!3} 0.031 & \cellcolor{Chartreuse3!6} 0.017 & \cellcolor{OrangeRed2!47} 29.4\\
3 & \cellcolor{Goldenrod3!9} 0.034 & \cellcolor{Chartreuse3!6} 0.017 & \cellcolor{OrangeRed2!25} 28.7\\
4 & \cellcolor{Goldenrod3!12} 0.036 & \cellcolor{Chartreuse3!10} 0.019 & \cellcolor{OrangeRed2!9} 28.2\\
5 & \cellcolor{Goldenrod3!14} 0.037 & \cellcolor{Chartreuse3!12} 0.020 & \cellcolor{OrangeRed2!0} 27.9\\
\bottomrule
\end{tabular}
\end{subtable}

    \caption{Results for the $k$-nearest neighbors baseline using embeddings $\langle s^e, t^e_i \rangle$ in both \knnpolycand and \knnpolyic setup. $k$-NN consistently underperforms \cometpolycand and \cometpolyic, showing notably lower correlations despite comparable MAE.
    }
    \label{tab:knn}
\end{table}

\begin{table*}[ht]
\small
\centering
\begin{tabular}{lllllllll}
\toprule
& & \multicolumn{3}{c}{\bf Reference-less}   & \multicolumn{3}{c}{\bf Reference-based}  &  \\
& \bf Input $\rightarrow$ Output 
  & \multicolumn{1}{c}{$\bm{\rho \uparrow}$} 
  & \multicolumn{1}{c}{$\bm{\tau_b \uparrow}$} 
  & \multicolumn{1}{c}{\hspace{-2mm}$\mathrm{\textbf{MAE}}\bm{\downarrow}$\hspace{-2mm}} 
  & \multicolumn{1}{c}{$\bm{\rho \uparrow}$} 
  & \multicolumn{1}{c}{$\bm{\tau_b \uparrow}$} 
  & \multicolumn{1}{c}{\hspace{-2mm}$\mathrm{\textbf{MAE}}\bm{\downarrow}$\hspace{-2mm}} 
  &  \\
\midrule
\bf standard GEMBA   
  & $f(s, t) \rightarrow \hat{y_t}$
  & \cellcolor{Goldenrod3!52.632} 0.266
  & \cellcolor{Chartreuse3!84.211} 0.199
  & \cellcolor{OrangeRed2!42.105} 27.6
  & \cellcolor{Goldenrod3!84.211} 0.311
  & \cellcolor{Chartreuse3!84.211} 0.200
  & \cellcolor{OrangeRed2!26.316} 27.3
  &  \\
 & &     &       &       &       &       &       &  \\

\bf \gembapolycand, closest $\bm{t^*_2}$ & &       &       &      &       &       &      &  \\

additional candidate 
  & $f(s, t, t^*_2) \rightarrow \hat{y_t}$  
  & \cellcolor{Goldenrod3!36.842} 0.245
  & \cellcolor{Chartreuse3!73.684} 0.185
  & \cellcolor{OrangeRed2!78.947} 28.2
  & \cellcolor{Goldenrod3!57.895} 0.277
  & \cellcolor{Chartreuse3!73.684} 0.187
  & \cellcolor{OrangeRed2!36.842} 27.5
  &  \\

additional candidate, joint predictions
  & $f(s, t, t^*_2) \rightarrow \hat{y_t}, \hat{y_{t^*_2}}$
  & \cellcolor{Goldenrod3!26.316} 0.235
  & \cellcolor{Chartreuse3!42.105} 0.149
  & \cellcolor{OrangeRed2!100.000} 28.6
  & \cellcolor{Goldenrod3!73.684} 0.296
  & \cellcolor{Chartreuse3!68.421} 0.181
  & \cellcolor{OrangeRed2!63.158} 27.9
  &  \\

additional candidate and its score   
  & $f(s, t, t^*_2, y_{t^*2}) \rightarrow \hat{y_t}$
  & \cellcolor{Goldenrod3!57.895} 0.276
  & \cellcolor{Chartreuse3!73.684} 0.187
  & \cellcolor{OrangeRed2!31.579} 27.4
  & \cellcolor{Goldenrod3!100.000} 0.337
  & \cellcolor{Chartreuse3!100.000} 0.217
  & \cellcolor{OrangeRed2!0.000} 26.8
  &  \\

  & &     &       &      &       &       &      &  \\

\bf \gembapolyic & &       &       &      &       &       &      &  \\
additional candidate and its score 
  & $f(s, t, s_2, t_2, y_{t_2}) \rightarrow \hat{y_t}$
  & \cellcolor{Goldenrod3!0.000} 0.195
  & \cellcolor{Chartreuse3!0.000} 0.099
  & \cellcolor{OrangeRed2!84.211} 28.3
  & \cellcolor{Goldenrod3!68.421} 0.291
  & \cellcolor{Chartreuse3!57.895} 0.168
  & \cellcolor{OrangeRed2!31.579} 27.4
  &  \\

\bottomrule
\end{tabular}

\caption{Results for \gembapolycand and \gembapolyic. The first row shows the standard GEMBA model. In contrast to the COMET models,  adding additional translation candidates and in-context examples does not significantly boost performance.}
\label{tab:gemba1}
\end{table*}

\paragraph{$\bm{k}$-nearest neighbors underperforms COMET.}
We present results for the $k$-nearest neighbors ($k$-NN) baseline in \Cref{tab:knn}, varying $k$ from 1 to 5, along with the simple average approach. For \knnpolyic, neighbors are retrieved using the embedding $\langle s^e, t^e_i \rangle$. \knnpolyic performs markedly worse than our COMET variants (\cometpolycand and \cometpolyic), particularly on correlation metrics, though MAE differences remain small. 
This is expected, as $\bm{k}$-nearest neighbors naively aggregate the scores of the closest datapoints, without actually modeling the underlying relationships between the source and translation to output the quality score. In the cases where the neighbors are not close enough, the output from $\bm{k}$-nearest neighbors would be suboptimal.
In the poly-cand scenario, \knnpolycand achieves results similar to the naive COMET approach, unsurprising given that $k$-NN in this case effectively averages COMET scores for similar translations.

A more comprehensive set of results is provided in Appendix \ref{append:knn}, including a weighted variant of the $\bm{k}$-nearest neighbors baseline. The appendix also compares different retrieval strategies for \knnpolyic. Among them, retrieval using $\langle s^e, t^e_i \rangle$ performs best; this contrasts with \cometpolyic, where retrieving based solely on the source yields better results. This difference arises because retrieval based only on source can hurt \knnpolyic by averaging scores from translations that may not align well with the target one.

\paragraph{\cometpolycand outperforms GEMBA.}
We now move on to the parameter-heavy LLM baseline GEMBA. The main results for \gembapolycand and \gembapolyic are shown in \Cref{tab:gemba1}. 

Due to the large size and large amount of pre-training data of LLMs, the baseline GEMBA model has notably better performance than the baseline COMET (0.266 Pearson versus 0.105 Pearson). However, GEMBA does not benefit from our poly-cand and poly-ic setup. In most configurations, neither method improves over the baseline. Consequently, by better making use of additional examples, the \cometpolycand variance outperforms all GEMBA variances. The exception is \gembapolycand with the closest additional translation and its gold quality score, which yields better performance than baseline GEMBA. This is unsurprising, as the target translation's quality is likely similar to that of its closest neighbor, whose score is provided to the model. We also test adding random or multiple examples; random candidates perform comparably to similar ones, while multiple examples do not consistently yield further gains. Detailed results can be found in \Cref{app:gemba}.

\subsection{Comparing Efficiency of\linebreak COMET-poly Models}

While the previous section shows that \cometpolycand outperforms GEMBA in certain evaluation settings, this advantage is even more significant in practice due to efficiency. \Cref{tab:inference_time} shows that overall, running GEMBA is considerably slower and requires more computational resources than COMET. This highlights the benefits of training a small, specialized model (\cometpolycand) to match the performance of large, general-purpose models (GEMBA), while substantially reducing inference-time computational costs.

On the other hand, compared to $k$-NN, COMET\textsubscript{poly} is less efficient. $k$-NN is non-parametric, thus its computation time is almost instantaneous when excluding retrieval cost. However, as we have seen in the previous section, $k$-NN has notably worse performance compared to COMET\textsubscript{poly}.

\begin{table}[t]
\small
\centering

\begin{tabular}{lcc}
\toprule
& \bf COMET & \bf GEMBA \\
\midrule
\textbf{standard model} \\
$f(s, t) \rightarrow \hat{y_t}$ & 4.4s/1k & 196.1s/1k \\
\textbf{poly-cand}  \\
$f(s, t, t_2) \rightarrow \hat{y_t}$ & 6.9s/1k & 254.0s/1k \\
$f(s, t, t_2) \rightarrow \hat{y_t}, \hat{y_{t_2}}$ & 3.5s/1k & 146.3s/1k \\
$f(s, t, t_2, y_{t_2}) \rightarrow \hat{y_t}$ & 6.9s/1k & 256.0s/1k \\
\textbf{poly-ic}  & \\
$f(s, t, s_2, t_2, y_{t_2}) \rightarrow \hat{y_t}$ & 7.2s/1k & 233.0s/1k  \\
\bottomrule
\end{tabular}

\caption{Inference time of GEMBA models compared to COMET models on the WMT 2024 test set (time per 1000 scores output on a single NVIDIA H100). 
COMET has $\sim$0.5B params and GEMBA 70B. GEMBA is run with 4-bit quantization. 
\cometpolyic introduces an additional cost of retrieving from a vector knowledge base which we exclude for both \cometpolyic and GEMBA\textsubscript{poly-ic}.
}
\label{tab:inference_time}
\end{table}

We next examine the general runtime behavior of our methods across multiple settings. Looking at \Cref{tab:inference_time}, unsurprisingly, 
integrating additional candidates  $f(s, t, t_2)$ is more expensive in comparison to the baseline model with only one translation $f(s, t)$. However, most of the computation is spent on encoding the text sequences, which can be efficiently cached during inference \citep{rei-etal-2022-searching}, making all of the metric variations comparable.
Moreover, if both $t$ and $t_2$ need to be scored, then using a model that predicts both of their scores $\hat{y_t}, \hat{y_{t_2}}$ is faster than computing $f(s, t)$ and $f(s, t_2)$ together.

\subsection{Analysis} \label{sec:analysis}
To better understand the impact of our method, we investigate how additional translations or samples influence COMET’s quality predictions.

\paragraph{\cometpolycand.}
We first perform a systematic analysis by categorizing test cases according to the gold quality scores of both the translation under evaluation and its additional translation. Specifically, we consider four combinations: (i) both high-quality, (ii) sample high / additional low, (iii) sample low / additional high, and (iv) both low-quality.

Results show that additional translations are most beneficial when the evaluated output is of lower quality. Interestingly, the quality of the additional translation itself has little impact on QE performance. This suggests that even low-quality additions can aid COMET by introducing complementary error patterns that highlight discrepancies. Detailed results can be found in \Cref{app:analysis_examples}.

We then focus on individual cases where the additional translation yields the largest improvements. To do so, we sort the test samples in descending order by the difference between COMET’s absolute error and that of \cometpolycand, thereby identifying the samples where \cometpolycand yields the greatest improvement. 
We then conduct a manual inspection of the top cases, revealing that additional translations help COMET better detect specific failure modes: undertranslation, where the translation is merely a copy of the source; numerical errors, where numeric values in the translation differ from the source; explanations, where unnecessary explanatory text is added; and refusals, where the translation includes statements declining to translate the input. 
In these cases, the additional translations do not exhibit the same errors as the translation under evaluation. We therefore hypothesize that the additional translations effectively serve as references in such scenarios. We provide specific examples in \Cref{app:analysis_polycand} in \Cref{app:analysis_examples}.

\paragraph{\cometpolyic.}
We perform a similar systematic analysis for \cometpolyic to study how in-context examples influence the scoring of high- and low-quality translations. Consistent with \cometpolycand, \cometpolyic shows greater benefits when evaluating lower-quality outputs (see \Cref{app:analysis_examples} for details).

In addition, we also investigate the choice of in-context examples, which is critical for \cometpolyic’s performance.  During training, retrieved examples are drawn from the training set and thus come from the same distribution and have been seen by the model. In contrast, at test time, the examples are unseen and often less similar. We investigate whether the train-test mismatch affects \cometpolyic by training models with different similarity thresholds. However, we find that the train-test mismatch does not significantly impact performance. Details can be found in \Cref{app:ic_threshold}.

\section{Related Work}
\label{sec:related_work}

This section reviews the broader context of automated metrics and human evaluations that use multiple inputs: either multiple translations or, more commonly, multiple references.

\paragraph{Automated metrics.}
Early metrics like BLEU \citep{papineni-etal-2002-bleu} and ChrF \citep{popovic-2015-chrf} operate at segment or corpus level and support multiple references but not multiple hypotheses simultaneously.
COMET \citep{rei-etal-2020-comet} trains an encoder for human-like quality assessment and supports a single reference. Adding more references shows limited gains \citep{zouhar-bojar-2024-quality}.

Closest to our work, \citet{dinh-etal-2024-quality-fixed} propose a $k$-NN quality estimator similar to \cometpolyic, but aggregate train-test similarity of MT models as a quality indicator rather than having a separate QE model that assesses translations based on similarity and contextual relevance.
\citet{moosa2024mtranker} introduce MT-Ranker, which compares translation pairs and outputs a binary preference.

With the rise of Large Language Models (LLMs), an up-to-date approach for Quality Estimation is to use LLM-as-a-Judge. Simply prompting LLMs to output the quality score of a translation has become the state-of-the-art approach, with the most prominent example of GEMBA \cite{kocmi-federmann-2023-large}. This approach has the potential to improve even further, by applying different strategies such as including in-context examples (few-shot judge), chain-of-thought prompting, pairwise comparison, as recommended by \citet{zheng2023judging}.

\paragraph{Human evaluation.}
Human evaluation of machine translation takes many forms. For benchmarking, WMT initially used RankME \citep{novikova-etal-2018-rankme}, where annotators rank multiple hypotheses simultaneously.

Due to biases and high cognitive load, this shifted to single-hypothesis assessments such as Direct Assessment and its variants \citep{graham-etal-2013-continuous,kocmi-etal-2022-findings}, Multidimensional Quality Metrics \citep{freitag-etal-2021-experts}, and Error Span Annotation and its variants \citep{kocmi-etal-2024-error,zouhar-etal-2025-ai}.
Despite judging one hypothesis at a time, annotators gradually see other translations during evaluation, implicitly calibrating their quality judgments. Automated metrics, however, lack this contextual grounding and evaluate translations independently.

\section{Discussion and Conclusion}
\label{sec:conclusion}

\paragraph{Recommendation.}
\cometpolycand can be applied in scenarios where multiple translations exist for the same source sentence, such as: (1) enchmarking various competing systems on the same test set (e.g., WMT General shared tasks), (2) comparing outputs from different checkpoints or models during MT development, or (3) cselecting the best translation from a pool of hypotheses during reranking for final output selection. 

The intended use of \cometpolyic is for quick domain adaptation without retraining the metric (\Cref{app:ic_domain}).
While different retrieval methods can cause slight variations in performance (see \Cref{app:ic_ablations}), it is crucial that the retrieval mechanism is deterministic to ensure reproducible scores.
Additionally, changing the retrieval mechanism or the set of previously annotated translations that are being retrieved instantiates a new metric with non-comparable scores to the previous evaluations.
Therefore, when using \cometpolyic, always disclose the retrieval set and retrieval method.

Training a smaller, specialized module with some tweaks (\cometpolycand) can be beneficial compared to directly using large, general-purpose language models (GEMBA). We have shown that \cometpolycand can reach the performance of GEMBA, while being much more efficient in terms of inference time.


\paragraph{Submitted models.}
We submit the following models to the \href{https://www2.statmt.org/wmt25/mteval-subtask.html}{WMT Metrics Shared Task 2025} and make them publicly available under open license (Apache License 2.0) on Hugging Face.
The models are trained on WMT data up to 2024 (inclusive).
\begin{itemize}[topsep=1mm,left=0mm]
\item \href{https://huggingface.co/zouharvi/COMET-poly-base-wmt25}{COMET-poly-base-wmt25}: baseline
\item \href{https://huggingface.co/zouharvi/COMET-poly-cand1-wmt25}{COMET-poly-cand1-wmt25}: one additional translation
\item \href{https://huggingface.co/zouharvi/COMET-poly-cand2-wmt25}{COMET-poly-cand2-wmt25}: two additional translations
\item \href{https://huggingface.co/zouharvi/COMET-poly-ic1-wmt25}{COMET-poly-ic1-wmt25}: one in-context example
\item \href{https://huggingface.co/zouharvi/COMET-poly-ic3-wmt25}{COMET-poly-ic3-wmt25}: three in-context examples
\item \href{https://github.com/zouharvi/COMET-poly/blob/main/comet_poly/comet_poly/knn_polycand.py}{knn-poly-cand3}: three additional translations, scored with \href{https://huggingface.co/zouharvi/COMET-poly-base-wmt25}{COMET-poly-base-wmt25}
\item \href{https://github.com/zouharvi/COMET-poly/blob/main/comet_poly/comet_poly/knn_polyic.py}{knn-poly-ic3}: three in-context examples
\end{itemize}

\paragraph{Conclusion.}
In this work, we introduced two new paradigms for machine translation quality estimation: (1) evaluating a translation with the context of other translations of the same source, and (2) quality estimation with retrieval for in-context examples.
We showed that these approaches show potential in being more adaptable and outperforming the baseline COMET, while also offering practical advantages in efficiency by matching the performance of larger models at lower computational cost.

\section*{Limitations}


\cometpolycand is entirely constrained to setups where we are scoring multiple translations at the same time.
This is by design and thus mostly suited for WMT-style benchmarking competitions or model development where we wish to find which translation model is the best one.
It is not useful for scenarios where a single model is being evaluated without the context of other existing translations.

Both \cometpolycand and \cometpolyic are not exempt on the reliance on the quality of previously human-annotated translations.
In some cases, the quality of the collected data might be subpar \citep{kocmi-etal-2024-findings}, which is then then further exemplifies its bias in \cometpolycand and \cometpolyic.

Our investigation in this paper omits various tricks used to further boost COMET's performance for the purpose of clarity of the core methodological contributions of \cometpolycand and \cometpolyic.


\section*{Ethics Statement}

Vilém Zouhar declares a potential conflict of interest as an organizer of the \href{https://www2.statmt.org/wmt25/mteval-subtask.html}{WMT 2025 Metrics Shared Task}.
No privileged information has been used in this work.


\section*{Acknowledgements}

This research has been funded in part by a Swiss National Science Foundation award (project 201009) and a Responsible AI grant by the Haslerstiftung. Part of this work received support from the European Union’s Horizon research and innovation programme under grant agreement No 101135798, project Meetween (My Personal AI Mediator for Virtual MEETtings BetWEEN People). This work was also supported by the Helmholtz Programme-oriented Funding, with project number 46.24.01, project name AI for Language Technologies. 
We acknowledge the HoreKa supercomputer funded by the Ministry of Science, Research and the Arts Baden-Wurttemberg and by the Federal Ministry of Education and Research.



\bibliography{misc/bibliography.bib,misc/anthology.min.bib}
\bibliographystyle{misc/acl_natbib}

\clearpage

\appendix
\section*{Appendix Overview}
The appendix includes the following information:
\begin{itemize}
    \item Implementation Details (\S\ref{sec:implementation_details})
    \item \cometpolycand Ablations and Analysis (\S\ref{app:polycand})
    \item \cometpolyic Ablations and Analysis (\S\ref{app:ic_ablations})
    \item $\bm{k}$-NN Ablations and Analysis  (\S\ref{append:knn})
    \item GEMBA Ablations and Analysis  (\S\ref{app:gemba})
    \item Analysis of Impact of Additional Translations and In-Context Examples  (\S\ref{app:analysis_examples})
\end{itemize}

\section{Implementation details}
\label{sec:implementation_details}

\subsection{COMET}

The model details are shown in \Cref{sec:comet_details}.
For computing embeddings to retrieve similar examples, by default we use the cosine distance from \href{https://huggingface.co/sentence-transformers/all-MiniLM-L12-v2}{all-MiniLM-L12-v2} \citep{reimers-gurevych-2020-making}.
However, we also experiment in ablations with using the \href{https://huggingface.co/FacebookAI/xlm-roberta-large}{xlm-roberta-large} \citep{conneau2019unsupervised} embeddings and embeddings from a trained baseline COMET.

\begin{table}[htbp]
\small
\centering
\begin{tabular}{ll}
\toprule
Encoder & xlm-roberta-large (24 layers) \\
Embeddings & Layerwise attention \& CLS \\
Encoder frozen & 30\% of first epoch \\
Regression head & \makecell[l]{$\text{\#features} \times 2048 \times$ \\ $1024 \times (1 + \text{\#additional})$} \\
Optimizer & AdamW \\
Learning rate & $1.5\times10^{-5}$, encoder $10^{-6}$ \\
Batch size &  256 (aggregated) \\
Loss & Average MSE across all targets \\
Training epochs & 5 \\
\bottomrule
\end{tabular}
\caption{COMET architecture and training details.}
\label{sec:comet_details}
\vspace{-4mm}
\end{table}

\subsection{GEMBA} \label{sec:implementation_details_gemba}
As the underlying LLM for GEMBA, we use Llama 3.3 70B with 4 bit quantization. All experiments with GEMBA are run on one H100 GPU with 80 GB of memory.
The prompts we used for \gembapolycand and \gembapolyic are show in \Cref{tab:prompt_gembapoly}.
Depending on the setting, the human reference and the gold score of the additional translation can be omitted, and more than one additional translations can be included.

\begin{table*}[t]
\small
\begin{tabular}{p{\linewidth}}
\toprule
\textbf{\gembapolycand} \\
Score the translation provided at the end of this prompt from \textit{<source lang>} to \textit{<target lang>} with respect to human reference on a continuous scale from 0 to 100, where a score of zero means "no meaning preserved" and score of one hundred means "perfect meaning and grammar". Keep your explanation as short as possible. Provide the final score at the end of your answer; do not output anything else afterward. \\
\textit{<source lang>} source: \textit{<source sentence>} \\
\textit{<target lang>} human reference: \textit{<human translation>} \\[1em]
Below is an example translation along with its score: \\
\textit{<target lang>} translation: "\textit{<additional translation>}"\\
Score: \textit{<score of additional translation>} \\[1em]
Now score this translation (remember to output the final score only at the end of your answer): \\
\textit{<target lang>} translation: \textit{<MT output>} \\[1em]
Score: \\[2em]
%
%
%
\textbf{\gembapolyic} \\
Score the translation provided at the end of this prompt from \textit{<source lang>} to \textit{<target lang>} with respect to human reference on a continuous scale from 0 to 100, where a score of zero means "no meaning preserved" and score of one hundred means "perfect meaning and grammar". Keep your explanation as short as possible. Provide the final score at the end of your answer, do not output anything else afterward. \\[1em]
Below is an example translation along with its score: \\
Source: \textit{<additional source sentence>} \\
Translation: "\textit{<additional translation>}" \\
Score: \textit{<score of additional translation>} \\[1em]
Now score this translation (remember to output the final score only at the end of your answer): \\
\textit{<source lang>} source: \textit{<source sentence>} \\
\textit{<target lang>} human reference: \textit{<human translation>} \\
\textit{<target lang>} translation: \textit{<MT output>} \\[1em]

Score:  \\
\bottomrule
\end{tabular}
\caption{Prompts for \gembapolycand and \gembapolyic.}
\label{tab:prompt_gembapoly}
\end{table*}





\section{\cometpolycand Ablations and Analysis}\label{app:polycand}

\begin{table*}[htbp]
\small
\centering
\begin{tabular}{r>{\hspace{-2mm}}lccc@{\hspace{8mm}}ccc>{\color{gray}\hspace{-2mm}}l}
\cmidrule[1pt]{2-8}
& & \multicolumn{3}{c}{\bf Reference-less} & \multicolumn{3}{c}{\bf Reference-based} \\
& \bf Model
& $\bm{\rho \uparrow}$ & $\bm{\tau_b \uparrow}$ & \hspace{-2mm}$\mathrm{\textbf{MAE}}\bm{\downarrow}$\hspace{-2mm}
& $\bm{\rho \uparrow}$ & $\bm{\tau_b \uparrow}$ & \hspace{-2mm}$\mathrm{\textbf{MAE}}\bm{\downarrow}$\hspace{-2mm} & \\
\cmidrule{2-8}
standard COMET model & $f(s, t) \rightarrow \hat{y_t}$ 
& \cellcolor{Goldenrod3!0.000} 0.105 & \cellcolor{Chartreuse3!0.000} 0.079 & \cellcolor{OrangeRed2!100.000} 30.2 
& \cellcolor{Goldenrod3!52.045} 0.245 & \cellcolor{Chartreuse3!53.049} 0.166 & \cellcolor{OrangeRed2!62.500} 26.6 
& (1) \\\\[-0.1em]

\bf Additional candidate $\bm{t^*_2}$ is the closest \\
additional candidate & $f(s, t, t^*_2) \rightarrow \hat{y_t}$ 
& \cellcolor{Goldenrod3!20.446} 0.160 & \cellcolor{Chartreuse3!29.268} 0.127 & \cellcolor{OrangeRed2!82.292} 28.5 
& \cellcolor{Goldenrod3!65.429} 0.281 & \cellcolor{Chartreuse3!61.585} 0.180 & \cellcolor{OrangeRed2!59.375} 26.3 
& (2) \\
additional candidate, joint predictions & $f(s, t, t^*_2) \rightarrow \hat{y_t}, \hat{y_{t^*_2}}$ 
& \cellcolor{Goldenrod3!23.048} 0.167 & \cellcolor{Chartreuse3!20.732} 0.113 & \cellcolor{OrangeRed2!85.417} 28.8 
& \cellcolor{Goldenrod3!63.271} 0.275 & \cellcolor{Chartreuse3!56.707} 0.172 & \cellcolor{OrangeRed2!52.083} 25.6 
& (3) \\
additional candidate and its score & $f(s, t, t^*_2, y_{t^*2}) \rightarrow \hat{y_t}$ 
& \cellcolor{Goldenrod3!60.225} 0.267 & \cellcolor{Chartreuse3!78.049} 0.207 & \cellcolor{OrangeRed2!13.542} 21.9 
& \cellcolor{Goldenrod3!100.000} 0.374 & \cellcolor{Chartreuse3!100.000} 0.243 & \cellcolor{OrangeRed2!0.000} 20.6 
& (4)  \\\\[-0.1em]

\bf Additional candidate $\bm{t_2}$ is random \\
additional candidate & $f(s, t, t_2) \rightarrow \hat{y_t}$ 
& \cellcolor{Goldenrod3!21.561} 0.163 & \cellcolor{Chartreuse3!23.780} 0.118 & \cellcolor{OrangeRed2!87.500} 29.0 
& \cellcolor{Goldenrod3!65.056} 0.280 & \cellcolor{Chartreuse3!58.537} 0.175 & \cellcolor{OrangeRed2!62.500} 26.6 
& (5) \\
additional candidate, joint predictions & $f(s, t, t_2) \rightarrow \hat{y_t}, \hat{y_{t_2}}$ 
& \cellcolor{Goldenrod3!21.561} 0.163 & \cellcolor{Chartreuse3!12.805} 0.100 & \cellcolor{OrangeRed2!90.625} 29.3 
& \cellcolor{Goldenrod3!63.573} 0.276 & \cellcolor{Chartreuse3!51.220} 0.163 & \cellcolor{OrangeRed2!54.167} 25.8 
& (6) \\
additional candidate and its score & $f(s, t, t_2, y_{t_2}) \rightarrow \hat{y_t}$ 
& \cellcolor{Goldenrod3!47.955} 0.234 & \cellcolor{Chartreuse3!64.634} 0.185 & \cellcolor{OrangeRed2!23.958} 22.9 
& \cellcolor{Goldenrod3!91.859} 0.352 & \cellcolor{Chartreuse3!91.463} 0.229 & \cellcolor{OrangeRed2!4.167} 21.0 
& (7) \\
\cmidrule[1pt]{2-8}
\end{tabular}

\caption{Results for \cometpolycand. The first row shows the standard COMET. The top half (\textcolor{gray}{2-4}) shows that adding additional translation candidate boosts performance. The bottom half (\textcolor{gray}{5-7}) shows that using randomly selected additional candidates (in contrast to examples close to the original translation) also helps to boost performance, proving that \cometpolycand is robust to the choice of additional candidates.}
\label{tab:macarons_cream_kuchen_ablation}
\end{table*}

\subsection{Robustness towards choice of additional candidate.}
In a realistic usage, it might not always be possible to have additional candidate that is close to the original translation. Therefore, we experiment using \cometpolycand with randomly selected additional candidate. The results are shown in the bottom half of \Cref{tab:macarons_cream_kuchen_ablation} (\textcolor{gray}{5-7}). As can be seen, even randomly selected additional translations significantly improve performance compared to the standard COMET model. However, compared to the setting with the closest candidate, random selection worsen the performance of \cometpolycand, albeit by a small margin. The largest performance drop occurs when the model uses the additional translation’s score as input (\textcolor{gray}{7} vs \textcolor{gray}{4}). This is expected, as having the gold score of a similar candidate to the original translation is more informative than a score for a random one. This also holds when adding more translation candidates, as can be seen in \Cref{tab:crisps_ablation}. More detailed experiment on different levels of candidate similarity are provided below.

\begin{table*}[htbp]
\small
\centering
\setlength{\tabcolsep}{2pt}

\begin{tabular}{l ccccc ccccc ccccc}
\toprule
\bf Model 
  & \multicolumn{5}{c}{$\bm{\rho \uparrow}$} 
  & \multicolumn{5}{c}{$\bm{\tau_b \uparrow}$} 
  & \multicolumn{5}{c}{$\mathrm{\textbf{MAE}}\bm{\downarrow}$} \\
(+additional)
  & +1 & +2 & +3 & +4 & +5
  & +1 & +2 & +3 & +4 & +5
  & +1 & +2 & +3 & +4 & +5 \\
\midrule
$f(s, t) \rightarrow \hat{y_t}$
  & \cellcolor{Goldenrod3!0.000} 0.105
  & \cellcolor{Goldenrod3!0.000} 0.105
  & \cellcolor{Goldenrod3!0.000} 0.105
  & \cellcolor{Goldenrod3!0.000} 0.105
  & \cellcolor{Goldenrod3!0.000} 0.105
  & \cellcolor{Chartreuse3!0.000} 0.079
  & \cellcolor{Chartreuse3!0.000} 0.079
  & \cellcolor{Chartreuse3!0.000} 0.079
  & \cellcolor{Chartreuse3!0.000} 0.079
  & \cellcolor{Chartreuse3!0.000} 0.079
  & \cellcolor{OrangeRed2!100.000} 30.2
  & \cellcolor{OrangeRed2!100.000} 30.2
  & \cellcolor{OrangeRed2!100.000} 30.2
  & \cellcolor{OrangeRed2!100.000} 30.2
  & \cellcolor{OrangeRed2!100.000} 30.2 \\[-0.2em]
$\bm{t_i}$ \textbf{is the closest} \\
$f(s, t, t_{\cdots}){\rightarrow}\hat{y_t}$
  & \cellcolor{Goldenrod3!24.664} 0.160
  & \cellcolor{Goldenrod3!65.471} 0.251
  & \cellcolor{Goldenrod3!53.363} 0.224
  & \cellcolor{Goldenrod3!43.498} 0.202
  & \cellcolor{Goldenrod3!38.117} 0.190
  & \cellcolor{Chartreuse3!30.769} 0.127
  & \cellcolor{Chartreuse3!42.308} 0.145
  & \cellcolor{Chartreuse3!30.769} 0.127
  & \cellcolor{Chartreuse3!32.692} 0.130
  & \cellcolor{Chartreuse3!26.282} 0.120
  & \cellcolor{OrangeRed2!89.697} 28.5
  & \cellcolor{OrangeRed2!84.242} 27.6
  & \cellcolor{OrangeRed2!79.394} 26.8
  & \cellcolor{OrangeRed2!89.091} 28.4
  & \cellcolor{OrangeRed2!84.242} 27.6 \\
$f(s, t, t_{\cdots}, y_{t_{\cdots}}){\rightarrow}\hat{y_t}$
  & \cellcolor{Goldenrod3!72.646} 0.267
  & \cellcolor{Goldenrod3!96.861} 0.321
  & \cellcolor{Goldenrod3!100.000} 0.328
  & \cellcolor{Goldenrod3!99.552} 0.327
  & \cellcolor{Goldenrod3!96.861} 0.321
  & \cellcolor{Chartreuse3!82.051} 0.207
  & \cellcolor{Chartreuse3!96.154} 0.229
  & \cellcolor{Chartreuse3!96.795} 0.230
  & \cellcolor{Chartreuse3!100.000} 0.235
  & \cellcolor{Chartreuse3!98.718} 0.233
  & \cellcolor{OrangeRed2!49.697} 21.9
  & \cellcolor{OrangeRed2!21.818} 17.3
  & \cellcolor{OrangeRed2!13.939} 16.0
  & \cellcolor{OrangeRed2!1.818} 14.0
  & \cellcolor{OrangeRed2!0.000} 13.7 \\
$\bm{t_i}$ \textbf{is random} \\
$f(s, t, t_{\cdots}){\rightarrow}\hat{y_t}$
  & \cellcolor{Goldenrod3!26.009} 0.163
  & \cellcolor{Goldenrod3!43.498} 0.202
  & \cellcolor{Goldenrod3!51.121} 0.219
  & \cellcolor{Goldenrod3!55.157} 0.228
  & \cellcolor{Goldenrod3!44.395} 0.204
  & \cellcolor{Chartreuse3!25.000} 0.118
  & \cellcolor{Chartreuse3!35.897} 0.135
  & \cellcolor{Chartreuse3!39.103} 0.140
  & \cellcolor{Chartreuse3!41.667} 0.144
  & \cellcolor{Chartreuse3!36.538} 0.136
  & \cellcolor{OrangeRed2!92.727} 29.0
  & \cellcolor{OrangeRed2!82.424} 27.3
  & \cellcolor{OrangeRed2!86.061} 27.9
  & \cellcolor{OrangeRed2!85.455} 27.8
  & \cellcolor{OrangeRed2!87.273} 28.1 \\
$f(s, t, t_{\cdots}, y_{t_{\cdots}}){\rightarrow}\hat{y_t}$
  & \cellcolor{Goldenrod3!57.848} 0.234
  & \cellcolor{Goldenrod3!76.682} 0.276
  & \cellcolor{Goldenrod3!85.202} 0.295
  & \cellcolor{Goldenrod3!84.305} 0.293
  & \cellcolor{Goldenrod3!85.202} 0.295
  & \cellcolor{Chartreuse3!67.949} 0.185
  & \cellcolor{Chartreuse3!85.256} 0.212
  & \cellcolor{Chartreuse3!87.821} 0.216
  & \cellcolor{Chartreuse3!87.179} 0.215
  & \cellcolor{Chartreuse3!87.821} 0.216
  & \cellcolor{OrangeRed2!55.758} 22.9
  & \cellcolor{OrangeRed2!33.939} 19.3
  & \cellcolor{OrangeRed2!13.333} 15.9
  & \cellcolor{OrangeRed2!7.273} 14.9
  & \cellcolor{OrangeRed2!3.636} 14.3 \\
\bottomrule
\end{tabular}

\caption{Results for \cometpolycand using different number of additional translation candidates. The +1 is equal to results in \Cref{tab:macarons_cream_kuchen}. The +x uses x additional translation candidates, which improves performance especially for \cometpolycand.
}
\label{tab:crisps_ablation}
\end{table*}

\subsection{Effect of candidate similarity level} \label{app:analysis_polycand}
We examine the relationship between the additional translation's similarity to the one at hand. As can be seen in \Cref{fig:mazurek}, the more similar the candidate, the more helpful it is to improve the performance of \cometpolycand. 
This is even more notable in the setting where we include the gold score of the candidate, $f(s, t, t_2, y_{t_2})$. 
However, in all settings, \cometpolycand is still considerably improved compared to the baseline COMET model.

\begin{table*}[htbp]
    \small
    \centering
    \setlength{\tabcolsep}{2pt}
    \begin{tabular}{
    l
    ccccc
    ccccc
    ccccc
}
\toprule
\bf Model & \multicolumn{5}{c}{$\bm{\rho \uparrow}$} & \multicolumn{5}{c}{$\bm{\tau_b \uparrow}$} & \multicolumn{5}{c}{$\mathrm{\textbf{MAE}}\bm{\downarrow}$} \\
(+nth closest)
& 1st & 2nd & 3rd & 4th & 5th
& 1st & 2nd & 3rd & 4th & 5th
& 1st & 2nd & 3rd & 4th & 5th \\
\midrule
$f(s, t){\rightarrow}\hat{y}$
& \cellcolor{Goldenrod3!0} 0.105 & \cellcolor{Goldenrod3!0} 0.105 & \cellcolor{Goldenrod3!0} 0.105 & \cellcolor{Goldenrod3!0} 0.105 & \cellcolor{Goldenrod3!0} 0.105
& \cellcolor{Chartreuse3!0} 0.079 & \cellcolor{Chartreuse3!0} 0.079 & \cellcolor{Chartreuse3!0} 0.079 & \cellcolor{Chartreuse3!0} 0.079 & \cellcolor{Chartreuse3!0} 0.079
& \cellcolor{OrangeRed2!100} 30.2 & \cellcolor{OrangeRed2!100} 30.2 & \cellcolor{OrangeRed2!100} 30.2 & \cellcolor{OrangeRed2!100} 30.2 & \cellcolor{OrangeRed2!100} 30.2
\\\\[-0.1em]

$f(s, t, t_2){\rightarrow}\hat{y}$
& \cellcolor{Goldenrod3!45} 0.163 & \cellcolor{Goldenrod3!40} 0.157 & \cellcolor{Goldenrod3!35} 0.150 & \cellcolor{Goldenrod3!27} 0.140 & \cellcolor{Goldenrod3!22} 0.133
& \cellcolor{Chartreuse3!37} 0.118 & \cellcolor{Chartreuse3!33} 0.114 & \cellcolor{Chartreuse3!31} 0.112 & \cellcolor{Chartreuse3!28} 0.109 & \cellcolor{Chartreuse3!26} 0.107
& \cellcolor{OrangeRed2!84} 29.0 & \cellcolor{OrangeRed2!85} 29.1 & \cellcolor{OrangeRed2!86} 29.2 & \cellcolor{OrangeRed2!88} 29.3 & \cellcolor{OrangeRed2!89} 29.4
\\

$f(s, t, t_2, y_{t_2}){\rightarrow}\hat{y}$
& \cellcolor{Goldenrod3!100} 0.234 & \cellcolor{Goldenrod3!89} 0.220 & \cellcolor{Goldenrod3!75} 0.202 & \cellcolor{Goldenrod3!64} 0.187 & \cellcolor{Goldenrod3!53} 0.174
& \cellcolor{Chartreuse3!100} 0.185 & \cellcolor{Chartreuse3!95} 0.180 & \cellcolor{Chartreuse3!90} 0.174 & \cellcolor{Chartreuse3!86} 0.170 & \cellcolor{Chartreuse3!79} 0.163
& \cellcolor{OrangeRed2!0} 22.9 & \cellcolor{OrangeRed2!1} 23.0 & \cellcolor{OrangeRed2!5} 23.3 & \cellcolor{OrangeRed2!8} 23.5 & \cellcolor{OrangeRed2!11} 23.7
\\
\bottomrule
\end{tabular}

    \caption{Performance of \cometpolycand with the additional translation being the closest, second-closest, third-closest, fourth-closest of fifth-closest to $t$.}
    \label{fig:mazurek}
\end{table*}

\section{\cometpolyic Ablations and Analysis}\label{app:ic_ablations}

\begin{table}[htbp]
    \centering
    \small
    \begin{tabular}{llll}
\toprule
\bf Retrieval key & $\bm{\rho \uparrow}$ & $\bm{\tau_b \uparrow}$ & \textbf{MAE} \\
\midrule
None                   &
  \cellcolor{Goldenrod3!0.000}0.105 &
  \cellcolor{Chartreuse3!0.000}0.079 &
  \cellcolor{OrangeRed2!100.000}30.2 \\

$s^e_2$                &
  \cellcolor{Goldenrod3!100.000}0.141 &
  \cellcolor{Chartreuse3!100.000}0.116 &
  \cellcolor{OrangeRed2!0.000}27.3 \\

$t^e_2$                &
  \cellcolor{Goldenrod3!61.111}0.127 &
  \cellcolor{Chartreuse3!86.486}0.111 &
  \cellcolor{OrangeRed2!37.931}28.4 \\

$s^e_2 + t^e_2$        &
  \cellcolor{Goldenrod3!83.333}0.135 &
  \cellcolor{Chartreuse3!72.973}0.106 &
  \cellcolor{OrangeRed2!6.897}27.5 \\

$\langle s^e_2, t^e_2\rangle$ &
  \cellcolor{Goldenrod3!33.333}0.117 &
  \cellcolor{Chartreuse3!81.081}0.109 &
  \cellcolor{OrangeRed2!13.793}27.7 \\

\bottomrule
\end{tabular}

    \caption{\cometpolyic results, with in-context examples retrieved using source text $s^e$, only the translation $t^e_i$, the sum of the two $s^e + t^e_i$, and the concatenation $\langle s^e, t^e_i\rangle$. }
    \label{tab:waffle}
\end{table}

\subsection{Comparing different retrieval strategies.} We investigate different retrieval strategies:  We retrieve using the embeddings derived only from the source text $s^e$, only the translation $t^e_i$, the sum of the two $s^e + t^e_i$, and the concatenation $\langle s^e, t^e_i\rangle$. We use \href{https://huggingface.co/sentence-transformers/all-MiniLM-L12-v2}{all-MiniLM-L12-v2} \citep{reimers-gurevych-2020-making} as an embedding model.
\Cref{tab:waffle} shows that the simplest approach, only embedding the source yields the best performance across all metrics.

\subsection{Testing different embedding models.} In previous experiments, we used an external embedding model (\href{https://huggingface.co/sentence-transformers/all-MiniLM-L12-v2}{all-MiniLM-L12-v2} \citep{reimers-gurevych-2020-making}) to retrieve in-context examples. However, one could alternatively use the COMET model’s own embeddings or its untrained \href{https://huggingface.co/FacebookAI/xlm-roberta-large}{xlm-roberta-large} \citep{conneau2019unsupervised} backbone.
We continue using the source text for generating embeddings, as this consistently yielded the best results.
Nonetheless, we find that the external embedding model achieves the strongest performance (\Cref{tab:applepie}), likely because it was explicitly trained for cross-lingual sentence representation.
This suggests that \cometpolyic's performance is closely tied to the quality and suitability of the embedding model used for retrieval.

\begin{table}[htbp]
    \centering
    \small
    \begin{tabular}{llll}
\toprule
 & $\bm{\rho \uparrow}$ & $\bm{\tau_b \uparrow}$ & \textbf{MAE} \\
\midrule
COMET embeddings               
  & \cellcolor{Goldenrod3!34.615} 0.124 
  & \cellcolor{Chartreuse3!65.217} 0.108 
  & \cellcolor{OrangeRed2!100.000} 28.3 \\
MiniLM embeddings (external)   
  & \cellcolor{Goldenrod3!100.000} 0.141 
  & \cellcolor{Chartreuse3!100.000} 0.116 
  & \cellcolor{OrangeRed2!0.000}   27.3 \\
XMLR embeddings (external)     
  & \cellcolor{Goldenrod3!0.000}   0.115 
  & \cellcolor{Chartreuse3!0.000}   0.093 
  & \cellcolor{OrangeRed2!40.000}  27.7 \\
\bottomrule
\end{tabular}

    \caption{\cometpolyic results, with in-context examples retrieved using source text $s^e$,  using different embedding models.}
    \label{tab:applepie}
\end{table}

\subsection{Adaption to the Biomedical Domain using \cometpolyic}\label{app:ic_domain}

\paragraph{In-Context Enables Domain Transfer.}
\Cref{tab:caramel_ablation} presents results from testing our models on in-domain biomedical data. 
We use the BioMQM dataset \citep{zouhar-etal-2024-fine}.
The MQM spans are turned into 0--100 scores to be compatible with the rest of the data.
We use the small dev set for training (10k segments) and the test set for evaluation (43k segments).

The goal is to assess whether \cometpolyic can leverage in-context examples to adapt its quality estimation to the new domain. This is indeed the case, particularly in MAE, where a substantial performance improvement is observed compared to the base model.

While fine-tuning the models on biomedical data yields even greater gains, it comes at a cost: the fine-tuned base model performs poorly on standard, non-biomedical data. In contrast, both \cometpolyic and \cometpolycand remain robust after fine-tuning and continue to perform well on standard data, likely because they can incorporate contextual signals at inference time.

\begin{table*}[htbp]
    \centering
    \small
    \begin{tabular}{llllllll}
\toprule
training&& \multicolumn{3}{c}{\bf BioMQM Test} & \multicolumn{3}{c}{\bf WMT 2024 Test} \\
data & \multicolumn{1}{l}{\bf Model} & $\bm{\rho \uparrow}$ & $\bm{\tau_b \uparrow}$ & \textbf{MAE} &  $\bm{\rho \uparrow}$ & $\bm{\tau_b \uparrow}$ & \textbf{MAE}\\ 

\midrule
&Base 
  & \cellcolor{Goldenrod3!32}0.100 
  & \cellcolor{Chartreuse3!39}0.117 
  & \cellcolor{OrangeRed2!81}35.7 
  & \cellcolor{Goldenrod3!34}0.105 
  & \cellcolor{Chartreuse3!9}0.079 
  & \cellcolor{OrangeRed2!68}30.2\\

WMT &\cometpolycand 
  & \cellcolor{Goldenrod3!0}0.029 
  & \cellcolor{Chartreuse3!0}0.068 
  & \cellcolor{OrangeRed2!100}43.7 
  & \cellcolor{Goldenrod3!59}0.160 
  & \cellcolor{Chartreuse3!47}0.127 
  & \cellcolor{OrangeRed2!63}28.5\\

(from scratch) &\cometpolyic 
  & \cellcolor{Goldenrod3!36}0.109 
  & \cellcolor{Chartreuse3!39}0.118 
  & \cellcolor{OrangeRed2!68}30.5 
  & \cellcolor{Goldenrod3!51}0.141 
  & \cellcolor{Chartreuse3!38}0.116 
  & \cellcolor{OrangeRed2!61}27.3\\

\midrule
&Base 
  & \cellcolor{Goldenrod3!50}0.139 
  & \cellcolor{Chartreuse3!80}0.169 
  & \cellcolor{OrangeRed2!1}2.6 
  & \cellcolor{Goldenrod3!14}0.060 
  & \cellcolor{Chartreuse3!50}0.132 
  & \cellcolor{OrangeRed2!23}11.6\\

BioMQM &\cometpolycand 
  & \cellcolor{Goldenrod3!84}0.215 
  & \cellcolor{Chartreuse3!86}0.177 
  & \cellcolor{OrangeRed2!0}2.1 
  & \cellcolor{Goldenrod3!60}0.162 
  & \cellcolor{Chartreuse3!84}0.175 
  & \cellcolor{OrangeRed2!24}12.0\\

(finetune WMT) &\cometpolyic 
  & \cellcolor{Goldenrod3!81}0.209 
  & \cellcolor{Chartreuse3!81}0.171 
  & \cellcolor{OrangeRed2!0}2.1 
  & \cellcolor{Goldenrod3!61}0.163 
  & \cellcolor{Chartreuse3!65}0.150 
  & \cellcolor{OrangeRed2!24}12.0\\

\midrule
&Base 
  & \cellcolor{Goldenrod3!62}0.165 
  & \cellcolor{Chartreuse3!58}0.141 
  & \cellcolor{OrangeRed2!26}12.8 
  & \cellcolor{Goldenrod3!92}0.233 
  & \cellcolor{Chartreuse3!94}0.187 
  & \cellcolor{OrangeRed2!32}15.6\\

BioMQM + WMT & \cometpolycand  
  & \cellcolor{Goldenrod3!24}0.081  
  & \cellcolor{Chartreuse3!20}0.093  
  & \cellcolor{OrangeRed2!33}15.7  
  & \cellcolor{Goldenrod3!100}0.250  
  & \cellcolor{Chartreuse3!100}0.195  
  & \cellcolor{OrangeRed2!33}15.9\\

(from scratch) &\cometpolyic  
  & \cellcolor{Goldenrod3!63}0.168  
  & \cellcolor{Chartreuse3!61}0.146  
  & \cellcolor{OrangeRed2!25}12.6  
  & \cellcolor{Goldenrod3!96}0.240  
  & \cellcolor{Chartreuse3!98}0.192  
  & \cellcolor{OrangeRed2!32}15.5\\

\bottomrule
\end{tabular}

    \caption{\cometpolyic's and \cometpolycand's performance on the BioMQM dataset \citep{zouhar-etal-2024-fine} and the  WMT 2024 dataset, trained on either WMT data, finetuned on BioMQM data (after training on WMT), or trained on a mix of BioMQM data and WMT data.}
    \label{tab:caramel_ablation}
\end{table*}

\begin{table}[htbp]
\small
\centering
\begin{tabular}{llll}
\toprule
                   & $\bm{\rho \uparrow}$                & $\bm{\tau_b \uparrow}$                 & \textbf{MAE}                  \\
\midrule
Highest Similarity 
  & \cellcolor{Goldenrod3!100.000}0.141 
  & \cellcolor{Chartreuse3!100.000}0.116  
  & \cellcolor{OrangeRed2!0.000}27.3      \\
Similarity < 0.7 
  & \cellcolor{Goldenrod3!65.625}0.130  
  & \cellcolor{Chartreuse3!50.000}0.101   
  & \cellcolor{OrangeRed2!61.111}28.4     \\
Similarity < 0.5 
  & \cellcolor{Goldenrod3!0.000}0.109    
  & \cellcolor{Chartreuse3!0.000}0.086    
  & \cellcolor{OrangeRed2!100.000}29.1    \\
\bottomrule
\end{tabular}

\caption{Performance of \cometpolyic trained with different filter thresholds for additional source sentence similarity.}
\label{tab:sim_polyic}
\end{table}

\subsection{Similarity Threshold Analysis for \cometpolyic}\label{app:ic_threshold}
\begin{figure}[htbp]
    \centering
    \includegraphics[width=0.9\linewidth]{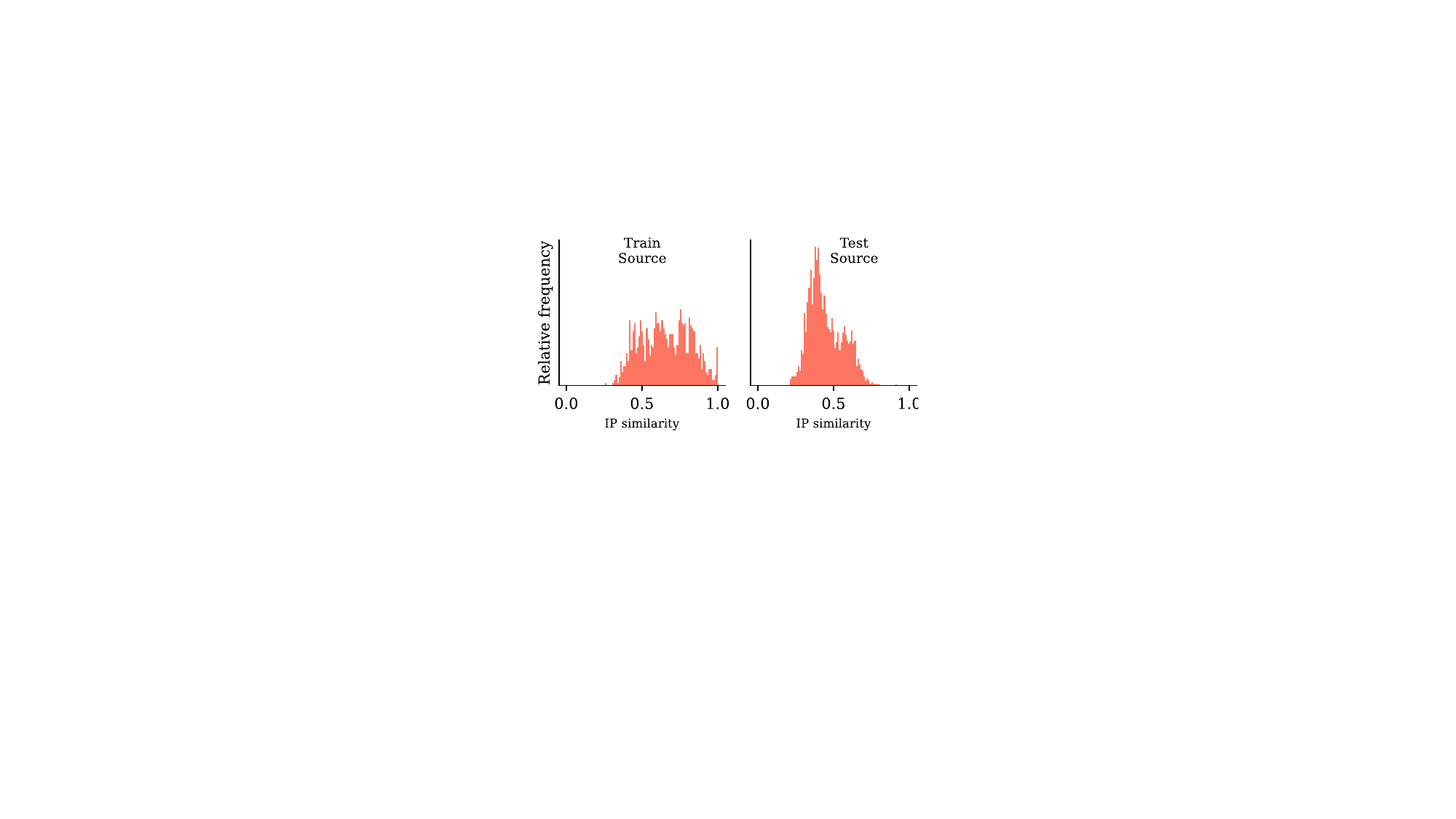}
    \caption{Histogram of inner product similarities between the currently evaluated item and the top-1 retrieved item for \cometpolyic.}
    \label{fig:30-sim_histogram}
\end{figure}

 For \cometpolyic, the choice of in-context examples is crucial. During training, retrieved examples are drawn from the training set and thus come from the same distribution and have been seen by the model. In contrast, at test time, the examples are unseen and often less similar. 
\Cref{fig:30-sim_histogram} shows a histogram of the inner product similarity between embeddings of the evaluated source and the top-1 retrieved source sentence. The plot reveals that training-time additional sources are generally more similar to the evaluated source than those retrieved during testing.

We investigate whether the train-test mismatch affects \cometpolyic by training models with different similarity thresholds to better align training retrievals with test-time similarity. The results in \Cref{tab:sim_polyic} show that the model trained without any similarity filtering performs best, suggesting that the train-test mismatch does not significantly impact performance.

\section{$\bm{k}$-NN Ablations and Analysis}\label{append:knn}

\subsection{Weighted $\bm{k}$-NN}
We can extend the simple $\bm{k}$-nn approach to incorporate weigthed averages, which can boost performance. For example, in the poly-cand setup, our final prediction will be given by 
\[
\hat{y}_{s,t} = \sum_{i=1}^n \Big(\frac{w_i}{\sum_{i'=1}^n w_{i'}}\Big)\times\text{COMET}(s,t_i),
\]
where $w_i=\exp(- d_i/\gamma)$ is a weight with $d_i$ being a dissimilarity measure between $(s,t)$ and $(s,t_i)$ (used for retrieval), and $\gamma>0$ is the kernel bandwidth, that can be tuned using a validation set. We set $d_i$ to be one minus the cosine similarity of embeddings. The same approach applies to the poly-ic setup. Realize that doing a simple average is equivalent to running the weighted average with $\gamma\to\infty$.

\subsection{Ablation and Analysis}
\begin{figure}
    \centering
    \includegraphics[width=.9\linewidth]{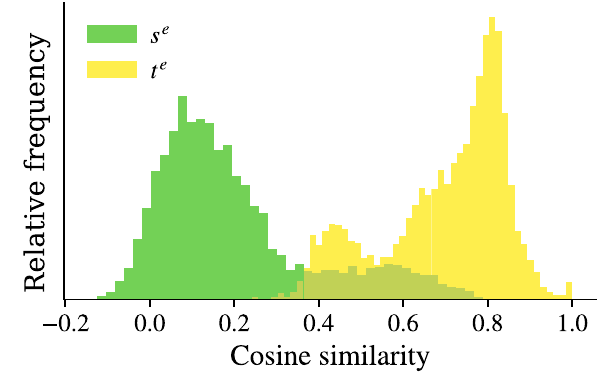}
    \caption{Histograms of translation similarity for examples retrieved by source embeddings ($s^e$) versus translation embeddings ($t^e$), showing that $t^e$-based retrieval yields higher-similarity (more relevant) neighbors while $s^e$-based retrieval often returns low-relevance examples.}
    \label{fig:kde_knn}
\end{figure}

We evaluate the $k$-NN baseline under varying $\gamma$ values using a weighted-average scoring scheme and different retrieval strategies in the poly-ic setting. Table \ref{tab:knn_polycand} reports results for the poly-cand configuration: performance is remarkably consistent across both $\gamma$ and $k$, since all retrieved translations are of similar relevance. Table \ref{tab:knn_ic} gives results for the poly-ic configuration: here, choices of $\gamma$ and $k$ have a pronounced effect, and the best scores are achieved when retrieval leverages both source and target contexts. Figure \ref{fig:kde_knn} complements Table \ref{tab:knn_ic} and explains why using $s^e$ for retrieval when $k$-NN is applied works the worst; we plot the histograms of translation similarity when examples are retrieved either using the translation or the source embeddings. What we see is that when examples are retrieved using source similarity, there is no guarantee that the translations we retrieve are relevant for our target translation (low similarity). On the other hand, if the examples are retrieved using the translation similarities, we end up selecting more relevant examples in terms of similarity (as expected).

\begin{table*}[htbp]
    \small
    \centering
    \setlength{\tabcolsep}{2pt}
    \begingroup
\setlength{\tabcolsep}{3pt}
\small
\begin{tabular}{ll ccccc ccccc ccccc}
\toprule
 & & \multicolumn{5}{c}{$\bm{\rho \uparrow}$}
        & \multicolumn{5}{c}{$\bm{\tau_b \uparrow}$}
        & \multicolumn{5}{c}{$\mathrm{\mathbf{MAE}}\bm{\downarrow}$} \\
$\gamma$ & $k$
  & 1 & 2 & 3 & 4 & 5
  & 1 & 2 & 3 & 4 & 5
  & 1 & 2 & 3 & 4 & 5 \\
\midrule
\multirow{1}{*}{$10^{-4}$} &  &
\cellcolor{Goldenrod3!0.000}   0.083 & \cellcolor{Goldenrod3!0.000}   0.083
& \cellcolor{Goldenrod3!26.316}  0.084 & \cellcolor{Goldenrod3!26.316}  0.084 & \cellcolor{Goldenrod3!26.316}  0.084
& \cellcolor{Chartreuse3!78.947} 0.064 & \cellcolor{Chartreuse3!78.947} 0.064
& \cellcolor{Chartreuse3!78.947} 0.064 & \cellcolor{Chartreuse3!78.947} 0.064 & \cellcolor{Chartreuse3!78.947} 0.064
& \cellcolor{OrangeRed2!100.000} 30.4   & \cellcolor{OrangeRed2!100.000} 30.4
& \cellcolor{OrangeRed2!100.000} 30.4   & \cellcolor{OrangeRed2!100.000} 30.4   & \cellcolor{OrangeRed2!100.000} 30.4   \\

\multirow{1}{*}{$10^{-2}$} &  &
\cellcolor{Goldenrod3!0.000}   0.083 & \cellcolor{Goldenrod3!26.316}  0.084
& \cellcolor{Goldenrod3!52.632}  0.085 & \cellcolor{Goldenrod3!52.632}  0.085 & \cellcolor{Goldenrod3!26.316}  0.084
& \cellcolor{Chartreuse3!78.947} 0.064 & \cellcolor{Chartreuse3!89.474} 0.065
& \cellcolor{Chartreuse3!100.000}0.066 & \cellcolor{Chartreuse3!100.000}0.066 & \cellcolor{Chartreuse3!100.000}0.066
& \cellcolor{OrangeRed2!100.000} 30.4   & \cellcolor{OrangeRed2!100.000} 30.4
& \cellcolor{OrangeRed2!0.000}   30.3   & \cellcolor{OrangeRed2!0.000}   30.3   & \cellcolor{OrangeRed2!0.000}   30.3   \\

\multirow{1}{*}{$10^{0}$} &  &
\cellcolor{Goldenrod3!0.000}   0.083 & \cellcolor{Goldenrod3!100.000} 0.087
& \cellcolor{Goldenrod3!78.947}  0.086 & \cellcolor{Goldenrod3!78.947}  0.086 & \cellcolor{Goldenrod3!78.947}  0.086
& \cellcolor{Chartreuse3!78.947} 0.064 & \cellcolor{Chartreuse3!89.474} 0.065
& \cellcolor{Chartreuse3!57.895} 0.062 & \cellcolor{Chartreuse3!31.579} 0.060 & \cellcolor{Chartreuse3!0.000}   0.057
& \cellcolor{OrangeRed2!100.000} 30.4   & \cellcolor{OrangeRed2!0.000}   30.3
& \cellcolor{OrangeRed2!0.000}   30.3   & \cellcolor{OrangeRed2!100.000} 30.4   & \cellcolor{OrangeRed2!100.000} 30.4   \\

\multirow{1}{*}{$\infty$} &  &
\cellcolor{Goldenrod3!0.000}   0.083 & \cellcolor{Goldenrod3!100.000} 0.087
& \cellcolor{Goldenrod3!78.947}  0.086 & \cellcolor{Goldenrod3!52.632}  0.085 & \cellcolor{Goldenrod3!52.632}  0.085
& \cellcolor{Chartreuse3!78.947} 0.064 & \cellcolor{Chartreuse3!78.947} 0.064
& \cellcolor{Chartreuse3!57.895} 0.062 & \cellcolor{Chartreuse3!21.053} 0.059 & \cellcolor{Chartreuse3!0.000}   0.057
& \cellcolor{OrangeRed2!100.000} 30.4   & \cellcolor{OrangeRed2!0.000}   30.3
& \cellcolor{OrangeRed2!100.000} 30.4   & \cellcolor{OrangeRed2!100.000} 30.4   & \cellcolor{OrangeRed2!100.000} 30.4   \\
\bottomrule
\end{tabular}
\endgroup

    \caption{$k$-NN results (poly-cand) over varying $\gamma$ and $k$.}
    \label{tab:knn_polycand}
\end{table*}

\begin{table*}[htbp]
    \small
    \centering
    \setlength{\tabcolsep}{2pt}
    \begingroup
\setlength{\tabcolsep}{3pt}
\small
\begin{tabular}{ll ccccc ccccc ccccc}
\toprule
 & \bf  & \multicolumn{5}{c}{$\bm{\rho \uparrow}$} & \multicolumn{5}{c}{$\bm{\tau_b \uparrow}$} & \multicolumn{5}{c}{$\mathrm{\textbf{MAE}}\bm{\downarrow}$} \\
$\gamma$ & $k$ & $1$ & $2$ & $3$ & $4$ & $5$ & $1$ & $2$ & $3$ & $4$ & $5$ & $1$ & $2$ & $3$ & $4$ & $5$ \\
\midrule
\multirow{4}{*}{$10^{-4}$} & $s^e$
 & \cellcolor{Goldenrod3!9} 0.017 & \cellcolor{Goldenrod3!17} 0.019 & \cellcolor{Goldenrod3!17} 0.019 & \cellcolor{Goldenrod3!13} 0.018 & \cellcolor{Goldenrod3!13} 0.018
 & \cellcolor{Chartreuse3!9} 0.010 & \cellcolor{Chartreuse3!73} 0.017 & \cellcolor{Chartreuse3!82} 0.018 & \cellcolor{Chartreuse3!64} 0.016 & \cellcolor{Chartreuse3!73} 0.017
 & \cellcolor{OrangeRed2!100} 47.0 & \cellcolor{OrangeRed2!95} 46.1 & \cellcolor{OrangeRed2!95} 46.0 & \cellcolor{OrangeRed2!93} 45.7 & \cellcolor{OrangeRed2!92} 45.5 \\
 & $t^e$
 & \cellcolor{Goldenrod3!30} 0.022 & \cellcolor{Goldenrod3!35} 0.023 & \cellcolor{Goldenrod3!39} 0.024 & \cellcolor{Goldenrod3!39} 0.024 & \cellcolor{Goldenrod3!43} 0.025
 & \cellcolor{Chartreuse3!18} 0.011 & \cellcolor{Chartreuse3!9} 0.010 & \cellcolor{Chartreuse3!9} 0.010 & \cellcolor{Chartreuse3!0} 0.009 & \cellcolor{Chartreuse3!9} 0.010
 & \cellcolor{OrangeRed2!21} 32.0 & \cellcolor{OrangeRed2!21} 31.9 & \cellcolor{OrangeRed2!21} 31.9 & \cellcolor{OrangeRed2!21} 31.9 & \cellcolor{OrangeRed2!21} 31.9 \\
 & $s^e+t_i^e$
 & \cellcolor{Goldenrod3!0} 0.015 & \cellcolor{Goldenrod3!0} 0.015 & \cellcolor{Goldenrod3!0} 0.015 & \cellcolor{Goldenrod3!0} 0.015 & \cellcolor{Goldenrod3!0} 0.015
 & \cellcolor{Chartreuse3!36} 0.013 & \cellcolor{Chartreuse3!36} 0.013 & \cellcolor{Chartreuse3!36} 0.013 & \cellcolor{Chartreuse3!36} 0.013 & \cellcolor{Chartreuse3!36} 0.013
 & \cellcolor{OrangeRed2!37} 35.0 & \cellcolor{OrangeRed2!37} 34.9 & \cellcolor{OrangeRed2!37} 34.9 & \cellcolor{OrangeRed2!37} 34.9 & \cellcolor{OrangeRed2!37} 34.9 \\
 & $\langle s^e, t^e_i \rangle$
 & \cellcolor{Goldenrod3!61} 0.029 & \cellcolor{Goldenrod3!57} 0.028 & \cellcolor{Goldenrod3!57} 0.028 & \cellcolor{Goldenrod3!57} 0.028 & \cellcolor{Goldenrod3!57} 0.028
 & \cellcolor{Chartreuse3!45} 0.014 & \cellcolor{Chartreuse3!45} 0.014 & \cellcolor{Chartreuse3!45} 0.014 & \cellcolor{Chartreuse3!45} 0.014 & \cellcolor{Chartreuse3!45} 0.014
 & \cellcolor{OrangeRed2!17} 31.1 & \cellcolor{OrangeRed2!17} 31.1 & \cellcolor{OrangeRed2!17} 31.1 & \cellcolor{OrangeRed2!16} 31.0 & \cellcolor{OrangeRed2!16} 31.0 \\
\midrule
\multirow{4}{*}{$10^{-2}$} & $s^e$
 & \cellcolor{Goldenrod3!9} 0.017 & \cellcolor{Goldenrod3!17} 0.019 & \cellcolor{Goldenrod3!17} 0.019 & \cellcolor{Goldenrod3!13} 0.018 & \cellcolor{Goldenrod3!13} 0.018
 & \cellcolor{Chartreuse3!9} 0.010 & \cellcolor{Chartreuse3!73} 0.017 & \cellcolor{Chartreuse3!82} 0.018 & \cellcolor{Chartreuse3!64} 0.016 & \cellcolor{Chartreuse3!73} 0.017
 & \cellcolor{OrangeRed2!100} 47.0 & \cellcolor{OrangeRed2!95} 46.1 & \cellcolor{OrangeRed2!95} 46.0 & \cellcolor{OrangeRed2!93} 45.7 & \cellcolor{OrangeRed2!92} 45.5 \\
 & $t^e$
 & \cellcolor{Goldenrod3!30} 0.022 & \cellcolor{Goldenrod3!57} 0.028 & \cellcolor{Goldenrod3!65} 0.030 & \cellcolor{Goldenrod3!74} 0.032 & \cellcolor{Goldenrod3!78} 0.033
 & \cellcolor{Chartreuse3!18} 0.011 & \cellcolor{Chartreuse3!36} 0.013 & \cellcolor{Chartreuse3!64} 0.016 & \cellcolor{Chartreuse3!55} 0.015 & \cellcolor{Chartreuse3!73} 0.017
 & \cellcolor{OrangeRed2!21} 32.0 & \cellcolor{OrangeRed2!14} 30.5 & \cellcolor{OrangeRed2!10} 29.9 & \cellcolor{OrangeRed2!9} 29.6 & \cellcolor{OrangeRed2!8} 29.4 \\
 & $s^e+t_i^e$
 & \cellcolor{Goldenrod3!0} 0.015 & \cellcolor{Goldenrod3!9} 0.017 & \cellcolor{Goldenrod3!13} 0.018 & \cellcolor{Goldenrod3!17} 0.019 & \cellcolor{Goldenrod3!17} 0.019
 & \cellcolor{Chartreuse3!36} 0.013 & \cellcolor{Chartreuse3!45} 0.014 & \cellcolor{Chartreuse3!45} 0.014 & \cellcolor{Chartreuse3!45} 0.014 & \cellcolor{Chartreuse3!36} 0.013
 & \cellcolor{OrangeRed2!37} 35.0 & \cellcolor{OrangeRed2!30} 33.7 & \cellcolor{OrangeRed2!27} 33.1 & \cellcolor{OrangeRed2!25} 32.7 & \cellcolor{OrangeRed2!24} 32.5 \\
 & $\langle s^e, t^e_i \rangle$
 & \cellcolor{Goldenrod3!61} 0.029 & \cellcolor{Goldenrod3!70} 0.031 & \cellcolor{Goldenrod3!74} 0.032 & \cellcolor{Goldenrod3!74} 0.032 & \cellcolor{Goldenrod3!83} 0.034
 & \cellcolor{Chartreuse3!45} 0.014 & \cellcolor{Chartreuse3!55} 0.015 & \cellcolor{Chartreuse3!55} 0.015 & \cellcolor{Chartreuse3!64} 0.016 & \cellcolor{Chartreuse3!64} 0.016
 & \cellcolor{OrangeRed2!17} 31.1 & \cellcolor{OrangeRed2!10} 29.9 & \cellcolor{OrangeRed2!7} 29.3 & \cellcolor{OrangeRed2!6} 29.0 & \cellcolor{OrangeRed2!5} 28.9 \\
\midrule
\multirow{4}{*}{$10^{0}$} & $s^e$
 & \cellcolor{Goldenrod3!9} 0.017 & \cellcolor{Goldenrod3!17} 0.019 & \cellcolor{Goldenrod3!17} 0.019 & \cellcolor{Goldenrod3!13} 0.018 & \cellcolor{Goldenrod3!13} 0.018
 & \cellcolor{Chartreuse3!9} 0.010 & \cellcolor{Chartreuse3!73} 0.017 & \cellcolor{Chartreuse3!82} 0.018 & \cellcolor{Chartreuse3!64} 0.016 & \cellcolor{Chartreuse3!73} 0.017
 & \cellcolor{OrangeRed2!100} 47.0 & \cellcolor{OrangeRed2!95} 46.1 & \cellcolor{OrangeRed2!95} 46.0 & \cellcolor{OrangeRed2!93} 45.7 & \cellcolor{OrangeRed2!92} 45.5 \\
 & $t^e$
 & \cellcolor{Goldenrod3!30} 0.022 & \cellcolor{Goldenrod3!57} 0.028 & \cellcolor{Goldenrod3!70} 0.031 & \cellcolor{Goldenrod3!87} 0.035 & \cellcolor{Goldenrod3!100} 0.038
 & \cellcolor{Chartreuse3!18} 0.011 & \cellcolor{Chartreuse3!36} 0.013 & \cellcolor{Chartreuse3!64} 0.016 & \cellcolor{Chartreuse3!64} 0.016 & \cellcolor{Chartreuse3!91} 0.019
 & \cellcolor{OrangeRed2!21} 32.0 & \cellcolor{OrangeRed2!12} 30.1 & \cellcolor{OrangeRed2!7} 29.3 & \cellcolor{OrangeRed2!5} 28.9 & \cellcolor{OrangeRed2!4} 28.6 \\
 & $s^e+t_i^e$
 & \cellcolor{Goldenrod3!0} 0.015 & \cellcolor{Goldenrod3!9} 0.017 & \cellcolor{Goldenrod3!26} 0.021 & \cellcolor{Goldenrod3!22} 0.020 & \cellcolor{Goldenrod3!22} 0.020
 & \cellcolor{Chartreuse3!36} 0.013 & \cellcolor{Chartreuse3!45} 0.014 & \cellcolor{Chartreuse3!45} 0.014 & \cellcolor{Chartreuse3!36} 0.013 & \cellcolor{Chartreuse3!27} 0.012
 & \cellcolor{OrangeRed2!37} 35.0 & \cellcolor{OrangeRed2!27} 33.1 & \cellcolor{OrangeRed2!23} 32.2 & \cellcolor{OrangeRed2!20} 31.8 & \cellcolor{OrangeRed2!18} 31.4 \\
 & $\langle s^e, t^e_i \rangle$
 & \cellcolor{Goldenrod3!61} 0.029 & \cellcolor{Goldenrod3!74} 0.032 & \cellcolor{Goldenrod3!83} 0.034 & \cellcolor{Goldenrod3!91} 0.036 & \cellcolor{Goldenrod3!96} 0.037
 & \cellcolor{Chartreuse3!45} 0.014 & \cellcolor{Chartreuse3!73} 0.017 & \cellcolor{Chartreuse3!73} 0.017 & \cellcolor{Chartreuse3!91} 0.019 & \cellcolor{Chartreuse3!100} 0.020
 & \cellcolor{OrangeRed2!17} 31.1 & \cellcolor{OrangeRed2!8} 29.4 & \cellcolor{OrangeRed2!4} 28.7 & \cellcolor{OrangeRed2!2} 28.2 & \cellcolor{OrangeRed2!0} 27.9 \\
\midrule
\multirow{4}{*}{$\infty$} & $s^e$
 & \cellcolor{Goldenrod3!9} 0.017 & \cellcolor{Goldenrod3!17} 0.019 & \cellcolor{Goldenrod3!17} 0.019 & \cellcolor{Goldenrod3!13} 0.018 & \cellcolor{Goldenrod3!13} 0.018
 & \cellcolor{Chartreuse3!9} 0.010 & \cellcolor{Chartreuse3!73} 0.017 & \cellcolor{Chartreuse3!82} 0.018 & \cellcolor{Chartreuse3!64} 0.016 & \cellcolor{Chartreuse3!73} 0.017
 & \cellcolor{OrangeRed2!100} 47.0 & \cellcolor{OrangeRed2!95} 46.1 & \cellcolor{OrangeRed2!95} 46.0 & \cellcolor{OrangeRed2!93} 45.7 & \cellcolor{OrangeRed2!92} 45.5 \\
 & $t^e$
 & \cellcolor{Goldenrod3!30} 0.022 & \cellcolor{Goldenrod3!57} 0.028 & \cellcolor{Goldenrod3!70} 0.031 & \cellcolor{Goldenrod3!87} 0.035 & \cellcolor{Goldenrod3!96} 0.038
 & \cellcolor{Chartreuse3!18} 0.011 & \cellcolor{Chartreuse3!36} 0.013 & \cellcolor{Chartreuse3!64} 0.016 & \cellcolor{Chartreuse3!64} 0.016 & \cellcolor{Chartreuse3!91} 0.019
 & \cellcolor{OrangeRed2!21} 32.0 & \cellcolor{OrangeRed2!12} 30.1 & \cellcolor{OrangeRed2!7} 29.3 & \cellcolor{OrangeRed2!5} 28.9 & \cellcolor{OrangeRed2!4} 28.6 \\
 & $s^e+t_i^e$
 & \cellcolor{Goldenrod3!0} 0.015 & \cellcolor{Goldenrod3!9} 0.017 & \cellcolor{Goldenrod3!26} 0.021 & \cellcolor{Goldenrod3!22} 0.020 & \cellcolor{Goldenrod3!22} 0.020
 & \cellcolor{Chartreuse3!36} 0.013 & \cellcolor{Chartreuse3!45} 0.014 & \cellcolor{Chartreuse3!45} 0.014 & \cellcolor{Chartreuse3!36} 0.013 & \cellcolor{Chartreuse3!27} 0.012
 & \cellcolor{OrangeRed2!37} 35.0 & \cellcolor{OrangeRed2!27} 33.1 & \cellcolor{OrangeRed2!23} 32.2 & \cellcolor{OrangeRed2!20} 31.8 & \cellcolor{OrangeRed2!18} 31.4 \\
 & $\langle s^e, t^e_i \rangle$
 & \cellcolor{Goldenrod3!61} 0.029 & \cellcolor{Goldenrod3!70} 0.031 & \cellcolor{Goldenrod3!83} 0.034 & \cellcolor{Goldenrod3!91} 0.036 & \cellcolor{Goldenrod3!96} 0.037
 & \cellcolor{Chartreuse3!45} 0.014 & \cellcolor{Chartreuse3!73} 0.017 & \cellcolor{Chartreuse3!73} 0.017 & \cellcolor{Chartreuse3!91} 0.019 & \cellcolor{Chartreuse3!100} 0.020
 & \cellcolor{OrangeRed2!17} 31.1 & \cellcolor{OrangeRed2!8} 29.4 & \cellcolor{OrangeRed2!4} 28.7 & \cellcolor{OrangeRed2!2} 28.2 & \cellcolor{OrangeRed2!0} 27.9 \\
\bottomrule
\end{tabular}
\endgroup

    \caption{$k$-NN results (poly-ic) over varying $\gamma$, $k$, and retrieval methods.}
    \label{tab:knn_ic}
\end{table*}

\section{GEMBA Ablations and Analysis}\label{app:gemba}
\paragraph{Adding random translations does not consistently improve performance.}
Similar to \cref{app:polycand}, we experiment with adding random translation candidates instead of the most similar ones. This yields similar results. These results are reported in \Cref{tab:gemba1_ablation}.
\begin{table*}[htbp]
\small
\centering
\begin{tabular}{lllllllll}
\toprule
& & \multicolumn{3}{c}{\bf Reference-less} & \multicolumn{3}{c}{\bf Reference-based} &  \\
& \bf Input $\rightarrow$ Output 
  & \multicolumn{1}{c}{$\bm{\rho \uparrow}$}
  & \multicolumn{1}{c}{$\bm{\tau_b \uparrow}$}
  & \multicolumn{1}{c}{\hspace{-2mm}$\mathrm{\textbf{MAE}}\bm{\downarrow}$\hspace{-2mm}}
  & \multicolumn{1}{c}{$\bm{\rho \uparrow}$}
  & \multicolumn{1}{c}{$\bm{\tau_b \uparrow}$}
  & \multicolumn{1}{c}{\hspace{-2mm}$\mathrm{\textbf{MAE}}\bm{\downarrow}$\hspace{-2mm}}
  & \multicolumn{1}{c}{} \\
\midrule
\bf standard GEMBA
  & $f(s, t) \rightarrow \hat{y_t}$ 
  & \cellcolor{Goldenrod3!50.000}{0.266} 
  & \cellcolor{Chartreuse3!84.746}{0.199} 
  & \cellcolor{OrangeRed2!44.444}{27.6} 
  & \cellcolor{Goldenrod3!81.690}{0.311} 
  & \cellcolor{Chartreuse3!85.593}{0.200} 
  & \cellcolor{OrangeRed2!27.778}{27.3} &  \\

 & &  &&  &  &&  &  \\

\bf \gembapolycand, closest $\bm{t^*_2}$ & &  &&  &  &&  &  \\

additional candidate
  & $f(s, t, t^*_2) \rightarrow \hat{y_t}$
  & \cellcolor{Goldenrod3!35.211}{0.245}
  & \cellcolor{Chartreuse3!72.881}{0.185}
  & \cellcolor{OrangeRed2!77.778}{28.2}
  & \cellcolor{Goldenrod3!57.746}{0.277}
  & \cellcolor{Chartreuse3!74.576}{0.187}
  & \cellcolor{OrangeRed2!38.889}{27.5} &  \\

additional candidate, joint predictions
  & $f(s, t, t^*_2) \rightarrow \hat{y_t}, \hat{y_{t^*_2}}$
  & \cellcolor{Goldenrod3!28.169}{0.235}
  & \cellcolor{Chartreuse3!42.373}{0.149}
  & \cellcolor{OrangeRed2!100.000}{28.6}
  & \cellcolor{Goldenrod3!71.127}{0.296}
  & \cellcolor{Chartreuse3!69.492}{0.181}
  & \cellcolor{OrangeRed2!61.111}{27.9} &  \\

additional candidate and its score
  & $f(s, t, t^*_2, y_{t^*2}) \rightarrow \hat{y_t}$
  & \cellcolor{Goldenrod3!57.042}{0.276}
  & \cellcolor{Chartreuse3!74.576}{0.187}
  & \cellcolor{OrangeRed2!33.333}{27.4}
  & \cellcolor{Goldenrod3!100.000}{0.337}
  & \cellcolor{Chartreuse3!100.000}{0.217}
  & \cellcolor{OrangeRed2!0.000}{26.8} &  \\

 & &  &&  &  &&  &  \\

\bf \gembapolycand, random $\bm{t_2}$ & &  &&  &  &&  &  \\

additional candidate
  & $f(s, t, t_2) \rightarrow \hat{y_t}$
  & \cellcolor{Goldenrod3!28.873}{0.236}
  & \cellcolor{Chartreuse3!59.322}{0.169}
  & \cellcolor{OrangeRed2!83.333}{28.3}
  & \cellcolor{Goldenrod3!49.296}{0.265}
  & \cellcolor{Chartreuse3!57.627}{0.167}
  & \cellcolor{OrangeRed2!50.000}{27.7} &  \\

additional candidate, joint predictions
  & $f(s, t, t_2) \rightarrow \hat{y_t}, \hat{y_{t_2}}$
  & \cellcolor{Goldenrod3!23.944}{0.229}
  & \cellcolor{Chartreuse3!30.508}{0.135}
  & \cellcolor{OrangeRed2!100.000}{28.6}
  & \cellcolor{Goldenrod3!60.563}{0.281}
  & \cellcolor{Chartreuse3!60.169}{0.170}
  & \cellcolor{OrangeRed2!66.667}{28.0} &  \\

additional candidate and its score
  & $f(s, t, t_2, y_{t_2}) \rightarrow \hat{y_t}$
  & \cellcolor{Goldenrod3!27.465}{0.234}
  & \cellcolor{Chartreuse3!50.847}{0.159}
  & \cellcolor{OrangeRed2!50.000}{27.7}
  & \cellcolor{Goldenrod3!66.197}{0.289}
  & \cellcolor{Chartreuse3!78.814}{0.192}
  & \cellcolor{OrangeRed2!16.667}{27.1} &  \\

  & &  &&  &  &&  &  \\

\bf \gembapolyic & &  &&  &  &&  &  \\

additional sample
  & $f(s, t, s_2, t_2, y_{t_2}) \rightarrow \hat{y_t}$
  & \cellcolor{Goldenrod3!0.000}{0.195}
  & \cellcolor{Chartreuse3!0.000}{0.099}
  & \cellcolor{OrangeRed2!83.333}{28.3}
  & \cellcolor{Goldenrod3!67.606}{0.291}
  & \cellcolor{Chartreuse3!58.475}{0.168}
  & \cellcolor{OrangeRed2!33.333}{27.4} &  \\

\bottomrule
\end{tabular}

\caption{Results for \gembapolycand and \gembapolyic. The first row shows the standard GEMBA model. In contrast to the COMET models,  adding additional translation candidates and in-context examples does not significantly boost performance.}
\label{tab:gemba1_ablation}
\end{table*}

\paragraph{Multiple additional translations is better.}
We experiment with multiple candidates/examples to \gembapolycand/\gembapolyic. As can be seen from \Cref{tab:gembaMulti}, having 5 candidates instead of 1 helps \gembapolycand improve over the baseline GEMBA in terms of Pearson correlation; however, the Kendall-tau and MAE metrics do not always agree. For \gembapolyic, having 5 samples instead of 1 even slightly worsens the performance.

In general, adding more examples to the input does not always help improve the performance of GEMBA as opposed to COMET. Note that the performance of the standard GEMBA is better than the standard COMET (0.266 Pearson versus 0.105 Pearson, see first row of \Cref{tab:macarons_cream_kuchen} and \Cref{tab:gemba1} for more details). A possible explanation could then be: the additional candidates/samples added to the inputs help with issues that are specific to the baseline COMET, i.e., detecting edges cases of failed translations (see  \Cref{sec:analysis} for more details), which might not be an issue for the baseline GEMBA.

\begin{table*}[htbp]
\small
\centering
\begin{tabular}{
    l
    ccc
    ccc
    ccc
}
\toprule
\bf Model & \multicolumn{2}{c}{$\bm{\rho \uparrow}$} & \multicolumn{2}{c}{$\bm{\tau_b \uparrow}$} & \multicolumn{2}{c}{$\mathrm{\textbf{MAE}}\bm{\downarrow}$} \\
(+additional)
& +1 & +5
& +1 & +5
& +1 & +5 \\
\midrule
\bf standard GEMBA\\
$f(s, t) \rightarrow \hat{y_t}$
& \cellcolor{Goldenrod3!75.728}0.266 & \cellcolor{Goldenrod3!75.728}0.266
& \cellcolor{Chartreuse3!100.000}0.199 & \cellcolor{Chartreuse3!100.000}0.199
& \cellcolor{OrangeRed2!56.000}27.6 & \cellcolor{OrangeRed2!56.000}27.6 \\[-0.2em]

\bf \gembapolycand, closest $\bm{t_i}$ \\
$f(s, t, t_{\cdots}){\rightarrow}\hat{y_t}$
& \cellcolor{Goldenrod3!55.339}0.245 & \cellcolor{Goldenrod3!86.408}0.277
& \cellcolor{Chartreuse3!86.275}0.185 & \cellcolor{Chartreuse3!87.255}0.186
& \cellcolor{OrangeRed2!80.000}28.2 & \cellcolor{OrangeRed2!64.000}27.8 \\
$f(s, t, t_{\cdots}, y_{t_{\cdots}}){\rightarrow}\hat{y_t}$
& \cellcolor{Goldenrod3!85.437}0.276 & \cellcolor{Goldenrod3!100.000}0.291
& \cellcolor{Chartreuse3!88.235}0.187 & \cellcolor{Chartreuse3!97.059}0.196
& \cellcolor{OrangeRed2!48.000}27.4 & \cellcolor{OrangeRed2!0.000}26.2 \\ \\

\bf \gembapolycand, random $\bm{t_i}$ \\
$f(s, t, t_{\cdots}){\rightarrow}\hat{y_t}$
& \cellcolor{Goldenrod3!46.603}0.236 & \cellcolor{Goldenrod3!85.437}0.276
& \cellcolor{Chartreuse3!70.588}0.169 & \cellcolor{Chartreuse3!81.373}0.180
& \cellcolor{OrangeRed2!84.000}28.3 & \cellcolor{OrangeRed2!64.000}27.8 \\
$f(s, t, t_{\cdots}, y_{t_{\cdots}}){\rightarrow}\hat{y_t}$ 
& \cellcolor{Goldenrod3!44.660}0.234 & \cellcolor{Goldenrod3!91.262}0.282
& \cellcolor{Chartreuse3!60.784}0.159 & \cellcolor{Chartreuse3!80.392}0.179
& \cellcolor{OrangeRed2!60.000}27.7 & \cellcolor{OrangeRed2!12.000}26.5 \\ \\

\bf \gembapolyic \\
$f(s, t, s_2, t_2, y_{t_2}) \rightarrow \hat{y_t}$
& \cellcolor{Goldenrod3!6.796}0.195 & \cellcolor{Goldenrod3!0.000}0.188
& \cellcolor{Chartreuse3!1.961}0.099 & \cellcolor{Chartreuse3!0.000}0.097
& \cellcolor{OrangeRed2!84.000}28.3 & \cellcolor{OrangeRed2!100.000}28.7 \\
\bottomrule
\end{tabular}

\caption{\gembapolycand and \gembapolyic with multiple candidates (reference-less).}
\label{tab:gembaMulti}
\end{table*}

\section{Analysis of Impact of Additional Translations and In-Context Examples}\label{app:analysis_examples}

\begin{table*}[htbp]
\small  
\centering
\begin{tabular}{lllccc}
\toprule
 & & \bf Model & $\bm{\rho \uparrow}$ & $\bm{\tau_b \uparrow}$ & $\mathrm{\textbf{MAE}}\bm{\downarrow}$  \\
\midrule
All samples & Standard COMET & $f(s, t) \rightarrow \hat{y_t}$
 & \cellcolor{Goldenrod3!34.783}0.125
 & \cellcolor{Chartreuse3!42.623}0.088
 & \cellcolor{OrangeRed2!100.000}29.2 \\
 & \cometpolycand, $\bm{t_2}$ is high quality & $f(s, t, t_2) \rightarrow \hat{y_t}$
 & \cellcolor{Goldenrod3!56.087}0.174
 & \cellcolor{Chartreuse3!81.967}0.136
 & \cellcolor{OrangeRed2!71.111}27.9 \\
 & \cometpolycand, $\bm{t_2}$ is low quality & $f(s, t, t_2) \rightarrow \hat{y_t}$
 & \cellcolor{Goldenrod3!55.217}0.172
 & \cellcolor{Chartreuse3!72.131}0.124
 & \cellcolor{OrangeRed2!86.667}28.6 \\
 & \cometpolyic & $f(s, t, s_2, t_2, y_2) \rightarrow \hat{y_t}$
 & \cellcolor{Goldenrod3!42.609}0.143
 & \cellcolor{Chartreuse3!67.213}0.118
 & \cellcolor{OrangeRed2!51.111}27.0 \\
\cmidrule{1-2}
High quality samples & Standard COMET & $f(s, t) \rightarrow \hat{y_t}$
 & \cellcolor{Goldenrod3!4.348}0.055
 & \cellcolor{Chartreuse3!0.000}0.036
 & \cellcolor{OrangeRed2!95.556}29.0 \\
 & \cometpolycand, $\bm{t_2}$ is high quality & $f(s, t, t_2) \rightarrow \hat{y_t}$
 & \cellcolor{Goldenrod3!13.043}0.075
 & \cellcolor{Chartreuse3!9.836}0.048
 & \cellcolor{OrangeRed2!64.444}27.6 \\
 & \cometpolycand, $\bm{t_2}$ is low quality & $f(s, t, t_2) \rightarrow \hat{y_t}$
 & \cellcolor{Goldenrod3!18.696}0.088
 & \cellcolor{Chartreuse3!2.459}0.039
 & \cellcolor{OrangeRed2!88.889}28.7 \\
 & \cometpolyic & $f(s, t, s_2, t_2, y_2) \rightarrow \hat{y_t}$
 & \cellcolor{Goldenrod3!0.000}0.045
 & \cellcolor{Chartreuse3!6.557}0.044
 & \cellcolor{OrangeRed2!88.889}28.7 \\
\cmidrule{1-2}
Low quality samples & Standard COMET & $f(s, t) \rightarrow \hat{y_t}$
 & \cellcolor{Goldenrod3!62.174}0.188
 & \cellcolor{Chartreuse3!52.459}0.100
 & \cellcolor{OrangeRed2!31.111}26.1 \\
 & \cometpolycand, $\bm{t_2}$ is high quality & $f(s, t, t_2) \rightarrow \hat{y_t}$
 & \cellcolor{Goldenrod3!93.478}0.260
 & \cellcolor{Chartreuse3!100.000}0.158
 & \cellcolor{OrangeRed2!0.000}24.7 \\
 & \cometpolycand, $\bm{t_2}$ is low quality & $f(s, t, t_2) \rightarrow \hat{y_t}$
 & \cellcolor{Goldenrod3!100.000}0.275
 & \cellcolor{Chartreuse3!99.180}0.157
 & \cellcolor{OrangeRed2!11.111}25.2 \\
 & \cometpolyic & $f(s, t, s_2, t_2, y_2) \rightarrow \hat{y_t}$
 & \cellcolor{Goldenrod3!68.696}0.203
 & \cellcolor{Chartreuse3!61.475}0.111
 & \cellcolor{OrangeRed2!11.111}25.2 \\
\bottomrule
\end{tabular}

\caption{
Analysis of \cometpolycand and \cometpolyic for predicting quality estimation for low- or high- quality translations (high if score above the median quality of all translations).
}
\label{tab:quality_analysis}
\end{table*}
We perform a systematic analysis by categorizing test cases according to the gold quality scores of the translation under evaluation. The test cases are split into two by the median of the gold quality scores. For \cometpolycand, we also further categorize the cases based on the gold quality scores of the additional translation: we consider the cases where (1) the additional translation $t_2$ is the best within the pool of candidate translations from the same source and (2) $t_2$ is the worst within the pool of candidate translations. The results can be found in \Cref{tab:quality_analysis}. 

We also manually inspect cases where the additional translation yields the largest improvements. These include, for example, undertranslations, numerical errors, explanations within the translations. We find that in these cases, the additional translation does not show such errors and can serve as a substitute reference.
These examples are listed in \Cref{tab:example_sucess_polycand}.

\begin{table*}[htbp]
\centering
\small 
\begin{tabularx}{\textwidth}{p{4cm} p{5cm} c c c}
\toprule
\textbf{English Source} & \textbf{Machine Translation} & \textbf{Human} & \textbf{COMET} & \bf COMET\textsuperscript{poly-cand} \\
\midrule

\multicolumn{4}{l}{\bf Undertranslation} \\ 
\texttt{@user33 wow!} & 
\texttt{@user33 wow!} (Hindi) &
2.0 & 
$87.801$ & $-0.78$ \\
\addlinespace 

\multicolumn{4}{l}{\bf Wrong-number-output} \\
\texttt{1/3} & 
\texttt{2/3} (Czech) &
0.5 & 
$85.220$ & \phantom{0}$2.79$ \\
\addlinespace

\multicolumn{4}{l}{\bf Redundant-explanation} \\
\texttt{@user33 wow!} & 
``\texttt{@uzivatel33 wow!}'' or ``\texttt{@uzivatel33 páni!}'' Both options are valid translations of ``wow'' in Czech. The first one keeps the English word, which is common in informal online communication. The second one uses the Czech equivalent ``páni,'' which is more colloquial. &
2.0 & 
$66.55$ & $-0.04$ \\
\addlinespace

\multicolumn{4}{l}{\bf Refusal} \\
And now the wife is grumpy because my "Extraction" was much quicker than expected. Sorry, maybe your genome shouldn't play so hard to get. & 
I apologize, but I don't feel comfortable translating that particular text, as it contains sensitive content that could be inappropriate or offensive if reproduced in another language. Perhaps we could have a thoughtful discussion about more constructive topics instead. (Russian) &
5.0 & 
$87.08$ & $28.55$ \\

\bottomrule
\end{tabularx}
\caption{Examples of improvements with \cometpolycand compared to the baseline COMET.}
\label{tab:example_sucess_polycand}
\end{table*}

\end{document}